\PassOptionsToPackage{unicode=true}{hyperref} % options for packages loaded elsewhere
\PassOptionsToPackage{hyphens}{url}
\documentclass[]{article}
\usepackage{lmodern}
\usepackage{amssymb,amsmath}
\usepackage{ifxetex,ifluatex}
\usepackage{fixltx2e} % provides \textsubscript
\ifnum 0\ifxetex 1\fi\ifluatex 1\fi=0 % if pdftex
  \usepackage[T1]{fontenc}
  \usepackage[utf8]{inputenc}
  \usepackage{textcomp} % provides euro and other symbols
\else % if luatex or xelatex
  \usepackage{unicode-math}
  \defaultfontfeatures{Ligatures=TeX,Scale=MatchLowercase}
\fi
\IfFileExists{upquote.sty}{\usepackage{upquote}}{}
\IfFileExists{microtype.sty}{%
\usepackage[]{microtype}
\UseMicrotypeSet[protrusion]{basicmath} % disable protrusion for tt fonts
}{}
\IfFileExists{parskip.sty}{%
\usepackage{parskip}
}{% else
\setlength{\parindent}{0pt}
\setlength{\parskip}{6pt plus 2pt minus 1pt}
}
\usepackage{hyperref}
\hypersetup{
            pdftitle={Phylogenetic signal in phonotactics},
            pdfauthor={Jayden L. Macklin-Cordes, Claire Bowern, Erich R. Round},
            pdfborder={0 0 0},
            breaklinks=true}
\usepackage[margin=1in]{geometry}
\usepackage{longtable,booktabs}
\IfFileExists{footnote.sty}{\usepackage{footnote}\makesavenoteenv{longtable}}{}
\usepackage{graphicx,grffile}
\makeatletter
\def\maxwidth{\ifdim\Gin@nat@width>\linewidth\linewidth\else\Gin@nat@width\fi}
\def\maxheight{\ifdim\Gin@nat@height>\textheight\textheight\else\Gin@nat@height\fi}
\makeatother
\setkeys{Gin}{width=\maxwidth,height=\maxheight,keepaspectratio}
\providecommand{\tightlist}{%
  \setlength{\itemsep}{0pt}\setlength{\parskip}{0pt}}
\ifx\paragraph\undefined\else
\let\oldparagraph\paragraph
\renewcommand{\paragraph}[1]{\oldparagraph{#1}\mbox{}}
\fi
\ifx\subparagraph\undefined\else
\let\oldsubparagraph\subparagraph
\renewcommand{\subparagraph}[1]{\oldsubparagraph{#1}\mbox{}}
\fi

\makeatletter
\def\fps@figure{htbp}
\makeatother

\usepackage{etoolbox}
\makeatletter
\providecommand{\subtitle}[1]{% add subtitle to \maketitle
  \apptocmd{\@title}{\par {\large #1 \par}}{}{}
}
\makeatother
\usepackage{booktabs}
\usepackage{longtable}
\usepackage{array}
\usepackage{multirow}
\usepackage{wrapfig}
\usepackage{float}
\usepackage{colortbl}
\usepackage{pdflscape}
\usepackage{tabu}
\usepackage{threeparttable}
\usepackage{threeparttablex}
\usepackage[normalem]{ulem}
\usepackage{makecell}
\usepackage{xcolor}

\title{Phylogenetic signal in phonotactics\footnote{This article has been submitted but not yet accepted for publication in a journal.}}
\author{Jayden L. Macklin-Cordes, Claire Bowern, Erich R. Round}
\date{}

\begin{document}
\maketitle

\hypertarget{abstract}{%
\section*{Abstract}\label{abstract}}
\addcontentsline{toc}{section}{Abstract}

Phylogenetic methods have broad potential in linguistics beyond tree inference. Here, we show how a phylogenetic approach opens the possibility of gaining historical insights from entirely new kinds of linguistic data---in this instance, statistical phonotactics. We extract phonotactic data from 111 Pama-Nyungan vocabularies and apply tests for \emph{phylogenetic signal}, quantifying the degree to which the data reflect phylogenetic history. We test three datasets: (1) binary variables recording the presence or absence of \emph{biphones} (two-segment sequences) in a lexicon (2) frequencies of transitions between segments, and (3) frequencies of transitions between natural sound classes. Australian languages have been characterized as having a high degree of phonotactic homogeneity. Nevertheless, we detect phylogenetic signal in all datasets. Phylogenetic signal is greater in finer-grained frequency data than in binary data, and greatest in natural-class-based data. These results demonstrate the viability of employing a new source of readily extractable data in historical and comparative linguistics.

\hypertarget{intro}{%
\section{Introduction}\label{intro}}

A defining methodological development in 21st century historical linguistics has been the adoption of computational phylogenetic methods for inferring phylogenetic trees of languages (Bowern \protect\hyperlink{ref-bowern_computational_2018}{2018}\protect\hyperlink{ref-bowern_computational_2018}{a}). The computational implementation of these methods means that it is possible to analyse large samples of languages, thereby inferring the phylogeny (evolutionary tree) of large language families at a scale and level of internal detail that would be difficult, if not impossible, to ascertain manually by a human researcher (Bowern \& Atkinson \protect\hyperlink{ref-bowern_computational_2012}{2012}: 827). There is more to phylogenetics than building trees, and there exists untapped potential to explore the language sciences and human history with a phylogenetic approach. For example, in linguistics, phylogenetic methods have been integrated with geography to infer population movements (Walker \& Ribeiro \protect\hyperlink{ref-walker_bayesian_2011}{2011}; Bouckaert, Bowern \& Atkinson \protect\hyperlink{ref-bouckaert_origin_2018}{2018}). In comparative biology, phylogenetic methods have been applied profitably to investigations of community ecology (Webb et al. \protect\hyperlink{ref-webb_phylogenies_2002}{2002}), ecological niche conservatism (Losos \protect\hyperlink{ref-losos_phylogenetic_2008}{2008}), paeleobiology (Sallan \& Friedman \protect\hyperlink{ref-sallan_heads_2012}{2012}) and quantitative genetics (Villemereuil \& Nakagawa \protect\hyperlink{ref-de_villemereuil_general_2014}{2014}). At the heart of these methods, however, is a sound understanding of the evolutionary dynamics of comparative structures. In this paper, we pesent a foundational step by detecting \emph{phylogenetic signal}, the tendency of related species (in our case, language varieties) to share greater-than-chance resemblances (Blomberg \& Garland \protect\hyperlink{ref-blomberg_tempo_2002}{2002}), in quantitative phonotactic variation.

Throughout recent advances in linguistic phylogenetics, less attention has been paid to methodological development at the stage of data preparation. Large-scale linguistic phylogenetic studies continue, by-and-large, to rely on lexical data which have been manually coded according to the principles of the \emph{comparative method} (as described by Meillet \protect\hyperlink{ref-meillet_methode_1925}{1925}; Campbell \protect\hyperlink{ref-campbell_historical_2004}{2004}; Weiss \protect\hyperlink{ref-weiss_comparative_2014}{2014})---the comparative method being the long-standing gold-standard of historical linguistic methodology (Chang et al. \protect\hyperlink{ref-chang_ancestry-constrained_2015}{2015}; Bouckaert, Bowern \& Atkinson \protect\hyperlink{ref-bouckaert_origin_2018}{2018}; Kolipakam et al. \protect\hyperlink{ref-kolipakam_bayesian_2018}{2018}). This article demonstrates that phonotactics can also present a source of historical information. We find that, for a sample of 111 Pama-Nyungan language varieties, collections of relatively simple and semi-automatically-extracted phonotactic variables (termed \emph{characters} throughout) contain phylogenetic signal. This has positive implications for the utility of such phonotactic data in linguistic phylogenetic inquiry, but also introduces methodological considerations for phonological typology.

In Sections \ref{intro}--\ref{phylo-sig}, we discuss the motivations for looking at phonotactics as a source of historical signal, and we give some broader scientific context that motivates the methodological approach we take later on. In Sections \ref{materials}--\ref{phy-sig-classes}, we present tests for phylogenetic signal in phonotactic characters extracted from wordlists for 111 Pama-Nyungan language varieties. Section \ref{materials} details the materials used and reference phylogeny. Section \ref{phy-sig-bin} tests for phylogenetic signal in binary characters that code the presence or absence of biphones (two-segment sequences) in each wordlist, capturing information on the permissibility of certain sequences in a language. Section \ref{phy-sig-cont} also tests for phylogenetic signal in biphones, but extracts a finer-grained level of variation by taking into account the relative frequencies of transitions between segments. Section \ref{phy-sig-classes} groups segments into natural sound classes and tests for phylogenetic signal in characters coding the relative frequencies of transitions between different classes. We finish in Sections \ref{discussion}--\ref{conclusion} with discussion of the limitations of the study design, implications of the results and directions for future research.

\hypertarget{motiv}{%
\subsection{Motivations}\label{motiv}}

There are at least two reasons why consideration of alternative data sources could be fruitful in historical linguistics. The first is that a bottleneck persists in linguistic phylogenetics when it comes to data processing. The data for most linguistic phylogenetic studies are lexical cognate data---typically binary characters marking the presence or absence of a cognate word in the lexicon of each language---which have been assembled from the manual judgements of expert linguists using the traditional comparative method of historical linguistics (e.g. Weiss \protect\hyperlink{ref-weiss_comparative_2014}{2014}). Although data assembled in this way is likely to remain the gold-standard in historical linguistics for the foreseeable future, it nevertheless constitutes slow and painstaking work (notwithstanding efforts to automate parts of the process; see List, Greenhill \& Gray \protect\hyperlink{ref-list_potential_2017}{2017}; Rama et al. \protect\hyperlink{ref-rama_are_2018}{2018}; List et al. \protect\hyperlink{ref-list_sequence_2018}{2018}). This restricts the pool of languages that can be included in phylogenetic research to those that have been more thoroughly documented, introducing the risk of a sampling bias, where relatively well-studied regions of the global linguistic landscape are over-represented in historical and comparative work.

The second motivation for considering alternative historical data sources is that there are inherent limitations associated with lexical data. Undetected semantic shifts and borrowed lexical items erode patterns of vertical inheritance in a language's lexicon. Put another way, these changes create noise in the historical signal of a language's lexicon. Chance resemblances between non-historically cognate words are another source of noise in lexical data. Eventually, semantic shifts, borrowings and chance resemblances will accumulate to a point where genuine historical signal is indistinguishable from noise. This imposes a maximal cap on the time-depth to which the comparative method can be applied, which is typically assumed to sit somewhere around 10,000 years BP, based on the approximate age of the Afro-Asiatic family (Nichols \protect\hyperlink{ref-nichols_sprung_1997}{1997}: 135). Some phylogenetic studies have attempted to push back the time-depth limitations of lexical data by using characters that code for a range of grammatical features, under the rationale that a language's grammatical structures should be more historically stable than its lexicon (Dunn et al. \protect\hyperlink{ref-dunn_structural_2005}{2005}; Rexová, Bastin \& Frynta \protect\hyperlink{ref-rexova_cladistic_2006}{2006}). However, contrary to expectation, a recent study suggests that grammatical characters evolve faster than lexical data (Greenhill et al. \protect\hyperlink{ref-greenhill_evolutionary_2017}{2017}). Differing rates of evolution are also found in phonology, specifically the rates of change in vowel inventories versus consonant inventories (Moran \& Verkerk \protect\hyperlink{ref-moran_differential_2018}{2018}; Moran, Grossman \& Verkerk \protect\hyperlink{ref-moran_investigating_2020}{2020}). An additional issue with grammatical characters is that the space of possibilities for a grammatical variable is often restricted. This means that chance similarities due to \emph{homoplasy} (parallel historical changes) will be much more frequent (c.f. Chang et al. \protect\hyperlink{ref-chang_ancestry-constrained_2015}{2015}). For example, many unrelated languages will share the same basic word order by chance, because there is a logical limit on the number of basic word order categories.

\hypertarget{why-phonotactics}{%
\subsection{Phonotactics as a source of historical signal}\label{why-phonotactics}}

The motivation for considering a language's phonotactics as a potential source of historical information is based partly on practical and partly on theoretical observations. From a practical perspective, it is possible to extract phonotactic data with relative ease, at scale, from otherwise resource-poor languages. This is because the bulk of a language's phonotactic system can be extracted directly from phonemicized wordlists. As long as there is a wordlist of suitable length (Dockum \& Bowern \protect\hyperlink{ref-dockum_swadesh_2019}{2019}) and a phonological analysis of the language, phonotactic information can be deduced and coded from the sequences of segments found in the wordlist with a high degree of automation. This modest minimum requirement with regards to language resources is a valuable property in less documented linguistic regions of the world. We detail the process of data extraction for this study in Section \ref{materials} below.

An additional benefit of extracting phonotactic data from wordlists is the potential for expanding the depth of comparative datasets. Although macro-scale studies, including hundreds or even thousands of the world's languages (in other words, \emph{broader} datasets), are increasingly common in comparative linguistics, less attention has been paid to the number of characters per language (dataset \emph{depth}). It is quite a different situation in evolutionary biology, where there has been tremendous growth in whole genome sequencing, thanks to technological advances and falling costs (Delsuc, Brinkmann \& Philippe \protect\hyperlink{ref-delsuc_phylogenomics_2005}{2005}; Wortley et al. \protect\hyperlink{ref-wortley_how_2005}{2005}). This, consequently, has led to tremendous growth in the depth of biological datasets. This is an important consideration because the quantity of characters required by modern computational phylogenetic methods can be substantial (Wortley et al. \protect\hyperlink{ref-wortley_how_2005}{2005}; Marin, Hedges \& Tamura \protect\hyperlink{ref-marin_undersampling_2018}{2018}). Certainly, phonotactic data is unlikely to approach the scale of large genomic datasets in biology, but it could effectively deepen historical linguistic datasets.

From a theoretical perspective, there is reason to suspect that the phonotactics of a language preserve a degree of historical signal. There is some evidence that when a borrowed word enters the lexicon of a language, speakers tend to adapt it to suit the phonotactic patterns of that language (Hyman \protect\hyperlink{ref-hyman_role_1970}{1970}; Silverman \protect\hyperlink{ref-silverman_multiple_1992}{1992}; Crawford \protect\hyperlink{ref-crawford_adaptation_2009}{2009}; Kang \protect\hyperlink{ref-kang_loanword_2011}{2011}). Consequently, in such a case the historical phonotactic structure of the lexicon remains largely intact, even as particular ancestral words are lost and replaced (a property termed \emph{pertinacity} by Dresher \& Lahiri \protect\hyperlink{ref-dresher_main_2005}{2005}). Similarly, historical phonotactic properties of a language will remain in the phonotactics of a language in the case of an undetected semantic shift.

Laboratory evidence shows that speakers have a high degree of sensitivity to the statistical distribution of phonological segments and structures when producing novel words. Examples of such studies include Coleman \& Pierrehumbert (\protect\hyperlink{ref-coleman_stochastic_1997}{1997}), Albright \& Hayes (\protect\hyperlink{ref-albright_rules_2003}{2003}) and Hayes \& Londe (\protect\hyperlink{ref-hayes_stochastic_2006}{2006}), among others (see Gordon \protect\hyperlink{ref-gordon_phonological_2016}{2016}: 20--21). Lexical innovation then, should have a relatively conservative impact on the frequency distributions of phonotactic characters. Every new word that enters a language's lexicon will have a minute impact on the frequencies of segments and particular sequences of segments in that language. But, over time, the cumulative effect of new lexicon entering a language on phonological and phonotactic frequency distributions will be more modest than if speakers generated new words with no regard for existing frequencies. Thus, there is reason to expect that quantitative phonotactic characters are likely to be conservative.

This is not to say that a language's phonotactic system remains completely immobile over time. Phonotactic systems are affected by sound changes and are not totally immune to borrowing. As mentioned above, frequencies of phonotactic characters will shift, however gradually, with the accumulation of lexical innovations. We make no strong claim about phonotactics being the key to a language's history. We merely note there are grounds to expect that phonotactic data will often be historically conservative, relative to cognate data which contains noise from lexical innovation, borrowing and semantic shift. Correspondingly, our hypothesis is that phonotactic data will contain relatively strong historical signal, which we test in Sections \ref{phy-sig-bin}--\ref{phy-sig-classes} below.

Many kinds of phonotactic structure exist, which could be studied phylogenetically. Here, because we wish to adhere to the basic methodological principle of studying maximally simple and clear cases first before progressing to more complex ones, we limit ourselves to the simplest of phonotactic structures, namely biphones. That being said, there is every reason to expect our results would generalize, perhaps with interesting variations, to other phonotactic structures. Moreover, many of those structures would have the same benefits as our biphones, in terms of their being readily generated in an automated fashion from wordlists. This will be a promising direction for future investigation.

\hypertarget{phylo-sig}{%
\section{Phylogenetic signal}\label{phylo-sig}}

The concept of \emph{phylogenetic signal} (Blomberg \& Garland \protect\hyperlink{ref-blomberg_tempo_2002}{2002}; Blomberg, Garland \& Ives \protect\hyperlink{ref-blomberg_testing_2003}{2003}: 717) originates in comparative biology, where it refers to the tendency of phylogenetically related species to resemble one another to a greater degree than would otherwise be expected by chance. This expectation derives from the evolutionary history shared between species. Two closely-related species, which share a relatively recent common ancestor, have had less time in which to diverge evolutionarily. We expect more distantly-related species, whose most recent common ancestor lies much further in the past, to tend to be more different, since they have spent longer on separate evolutionary paths.

Phylogenetic signal manifests itself as \emph{phylogenetic autocorrelation} in comparative studies. That is, species observations in a comparative dataset tend not to behave as independent data points, but rather pattern as a function of the amount of shared evolutionary history between species. For many statistical methods that assume data are independent and identically distributed (i.i.d.), this is a problem. Phylogenetic autocorrelation has long been recognized as an issue in linguistic typology and comparative biology, and both fields share comparable histories of developing sampling methodologies that attempt to correct for or offset phylogenetic relatedness in some way. More recent times have seen the rise of \emph{phylogenetic comparative methods}, statistical methods that directly account for phylogenetic autocorrelation, rather than offsetting it, beginning with foundational works by Felsenstein (\protect\hyperlink{ref-felsenstein_phylogenies_1985}{1985}) and Grafen (\protect\hyperlink{ref-grafen_phylogenetic_1989}{1989})\footnote{See Nunn (\protect\hyperlink{ref-nunn_comparative_2011}{2011}) for discussion.}. Although now practically ubiquitous in comparative biology, uptake of phylogenetic comparative methods has been slower in comparative linguistics (notwithstanding studies such as Dunn et al. \protect\hyperlink{ref-dunn_evolved_2011}{2011}; Maurits \& Griffiths \protect\hyperlink{ref-maurits_tracing_2014}{2014}; Verkerk \protect\hyperlink{ref-verkerk_diachronic_2014}{2014}; Birchall \protect\hyperlink{ref-birchall_comparison_2015}{2015}; Zhou \& Bowern \protect\hyperlink{ref-zhou_quantifying_2015}{2015}; Calude \& Verkerk \protect\hyperlink{ref-calude_typology_2016}{2016}; Dunn et al. \protect\hyperlink{ref-dunn_dative_2017}{2017}; Verkerk \protect\hyperlink{ref-verkerk_phylogenetic_2017}{2017}; Widmer et al. \protect\hyperlink{ref-widmer_np_2017}{2017}; Blasi et al. \protect\hyperlink{ref-blasi_human_2019}{2019}).

Since the turn of the century, methods have been developed for explicitly quantifying the degree of phylogenetic signal in a dataset (Revell et al. \protect\hyperlink{ref-revell_phylogenetic_2008}{2008}: 591). Measuring phylogenetic signal can be the first step of a comparative study, to test whether there is sufficient phylogenetic signal to necessitate implementation of a phylogenetic comparative method in a later stage of analysis, or to establish the suitability of standard statistical methods if no phylogenetic signal is detected. Measures of phylogenetic signal can also be used to re-evaluate the validity of older results that pre-date modern phylogenetic comparative methods, as in Freckleton, Harvey \& Pagel (\protect\hyperlink{ref-freckleton_phylogenetic_2002}{2002}). In other instances, the presence or absence of phylogenetic signal in certain data may be an interesting result in itself. In this study, we present a novel source of linguistic data which traditionally has not been considered a salient source of historical signal for historical linguistic study (indeed, given descriptions of Australian languages, it may have been considered a particularly unlikely source of historical signal; see Section \ref{sample}). We use measures of phylogenetic signal to test the hypothesis that our data contain historical information and, therefore, could contribute to future historical linguistic study.

Blomberg, Garland \& Ives (\protect\hyperlink{ref-blomberg_testing_2003}{2003}) provide a set of statistics for measuring phylogenetic signal, which remains prevalent today (for example, Balisi, Casey \& Valkenburgh \protect\hyperlink{ref-balisi_dietary_2018}{2018}; Hutchinson, Gaiarsa \& Stouffer \protect\hyperlink{ref-hutchinson_contemporary_2018}{2018}; Leff et al. \protect\hyperlink{ref-leff_predicting_2018}{2018}). We use one of these statistics, \(K\). The \(K\) statistic has the desirable property of being independent of the size and shape of the phylogenetic tree being investigated, which means that studies with different sample sizes can be compared directly. Briefly (following Blomberg, Garland \& Ives \protect\hyperlink{ref-blomberg_testing_2003}{2003}: 722), the calculation of \(K\) requires three components: (i) character data (i.e., observations for the variable of interest); (ii) a \emph{reference phylogeny}, a phylogenetic tree which has been generated independently from the character data; and (iii) a \emph{Brownian motion} model of evolution.\footnote{A \emph{Brownian motion} model of evolution describes a model of character evolution where the character can move up or down with equal probability as it evolves through time. Under this model of evolution, variance in character values throughout a phylogeny will increase proportionally as time elapses.} These components entail two assumptions of the method: the assumption that the reference phylogeny is an accurate representation of the phylogenetic history of the populations being studied and the assumption that Brownian motion accurately models the evolution of the character data. In practice, the reference phylogeny will be subject to uncertainty. We return to this point in Section \ref{discussion} and evaluate the robustness of our results against phylogenetic uncertainty. Similarly, in practice, the Brownian motion model may not be realistic. Nevertheless, it is a simple model and straightforward to implement, and thus commonly used as a starting point before exploring more complex models of evolution later on. We discuss this further in Section \ref{discussion} and outline possible extensions to the model for future study, taking sound change processes into account. To the extent that Brownian motion fails to model the evolution of phonotactic characters, this should make it more difficult to detect phylogenetic signal.

The \(K\) statistic is then calculated by, firstly, taking the mean squared error of the data (\(MSE_0\)), as measured from a \emph{phylogenetic mean}\footnote{Simply taking the mean of some variable would be misleading in cases where members of a particularly large clade happen to share similar values at an extreme end of the range. A \emph{phylogenetic mean} is an estimate of the mean which takes into account any overrepresentation by larger subclades (see, for example, Garland \& Díaz-Uriarte \protect\hyperlink{ref-garland_jr._polytomies_1999}{1999}).}, and dividing it by the mean squared error of the data (\(MSE\)), calculated using a variance-covariance matrix of phylogenetic distances between tips in the reference tree (see Blomberg, Garland \& Ives \protect\hyperlink{ref-blomberg_testing_2003}{2003} for a complete formula). This latter value, \(MSE\), will be small when the pattern of covariance in the data matches what would be expected given the phylogenetic distances in the reference tree, leading to a high \(MSE_0/MSE\) ratio and vice versa. Thus, a high \(MSE_0/MSE\) ratio indicates higher phylogenetic signal. Finally, the observed \(MSE_0/MSE\) ratio can be scaled according to the expected \(MSE_0/MSE\) ratio given a Brownian motion model of evolution. This gives a statistic, \(K\), which can be compared directly between studies using different trees. When \(K = 1\), this suggests a perfect match between the covariance observed in the data and what would be expected given the reference tree and the assumption of Brownian motion evolution. When \(K < 1\), close relatives in the tree bear less resemblance in the data than would be expected under the Brownian motion assumption. \(K > 1\) is also possible---this occurs where there is less variance in the data than expected, given the Brownian motion assumption and divergence times suggested by the reference tree. In other words, close relatives bear closer resemblance than would be expected if the variable evolved along the tree following a Brownian motion model of evolution.

Blomberg, Garland \& Ives (\protect\hyperlink{ref-blomberg_testing_2003}{2003}) also present a \emph{randomisation procedure} for testing whether the degree of phylogenetic signal in a dataset is statistically significant. The randomisation procedure utilizes Felsenstein's (1985) \emph{phylogenetic independent contrasts} (PICs) method. Felsenstein's insight is that, although two character values (\(x\) and \(y\)) from two sister taxa cannot be considered independent due to phylogenetic autocorrelation, the contrast between them (\(x - y\)) is phylogenetically independent, since these values can only diverge in the time since the two sisters split from their most recent common ancestor. Given a set of character data and a phylogenetic tree, Felsenstein (\protect\hyperlink{ref-felsenstein_phylogenies_1985}{1985}) presents a method for harvesting a whole set of phylogenetically independent data points, PICs, which can be used for statistical analysis in lieu of the raw set of observations. Blomberg, Garland \& Ives (\protect\hyperlink{ref-blomberg_testing_2003}{2003}) take advantage of the expectation that, given a Brownian motion model of evolution, PIC variance is expected to be proportional to time. PICs among more closely-related taxa will tend to be lower than more distant relatives, since they have had less time to diverge from common ancestors. The randomisation procedure first extracts PICs for a given character and records the variance. Then, it extracts PICs and records the variance after randomly shuffling character data among taxa (thereby destroying phylogenetic signal). PIC variance is recorded typically for many thousands of such random permutations. If the true PIC variance (for original, unshuffled data) is lower than the variance of PICs in \textgreater{} 95\% of random permutations, the null hypothesis of no phylogenetic signal can be rejected at the conventional 95\% confidence level.

In this study, we also use a second statistic, \(D\), which was developed to measure phylogenetic signal in binary data. The \(D\) statistic is described by Fritz \& Purvis (\protect\hyperlink{ref-fritz_selectivity_2010}{2010}). To summarize briefly, the \(D\) statistic is based on the sum of differences between sister tips and sister clades, \(\Sigma d\). First, differences between values at the tips of the tree are summed. Since \(D\) concerns binary variables, each taxon will either have a 0 or 1 value. At the level of the tips, then, all sister tips will either share the same value (in which case, the difference = 0) or one tip will have a 0 value and the other will have a 1 value (in which case, the difference = 1). Nodes immediately above the tree tips are given the average value of their daughter tips below (which, in a fully bifurcating phylogeny, will either be 0, 0.5 or 1). This process is repeated for all nodes in the tree, until a total sum of differences, \(\Sigma d\), is reached. At two extremes, data may be maximally clumped, such that all 1s are grouped together in the same clade in the tree and likewise for all 0s, or data may be maximally dispersed, such that no two sister tips share the same value (every pair of sisters contains a 1 and a 0, leading to a maximal sum of differences). Lying somewhere in between will be both (i) a distribution that is entirely random relative to phylogenetic structure and (ii) a distribution that is clumped exactly to the degree expected if the character evolved along the tree following a Brownian motion model of evolution. Two permutation procedures are used to determine where these two points lie for a given dataset and phylogenetic tree. Firstly, like Blomberg \emph{et al.}'s permutation test described above, character values are shuffled at random among tips of the tree many times over, thereby destroying phylogenetic signal. The sums of differences are taken from each random permutation to obtain a distribution of sums of differences, given phylogenetic randomness: \(\Sigma d_r\). Then, to obtain a contrasting distribution of sums of differences, the process of character evolution along the tree following a Brownian motion model is simulated many times over. Since Brownian motion is a model of evolution of continuous characters, and what we need here is a distribution of binary character values, the permutation test simulates the evolution of a continuous-valued character and then simply binarizes the tip values to 0 or 1 by observing whether they fall above or below a threshold value. This threshold is set to whatever level will produce the same proportion of 1s and 0s as observed in the real data. The sums of differences are then taken from each simulation, giving a distribution where phylogenetic signal is present: \(\Sigma d_b\). Finally, the \(D\) statistic is determined by scaling the observed sum of differences relative to the means of the two reference distributions just described:

\begin{equation}
D = \frac{\Sigma d_{obs} - mean\left( \Sigma d_{b} \right)}{mean\left( \Sigma d_{r} \right) - mean\left( \Sigma d_{b} \right)}
\end{equation}

Scaling \(D\) in this way provides a standardized statistic with the desirable property that it can be compared between different sets of data, with trees of different sizes and shapes, as with \(K\) for continuous characters. One disadvantage of \(D\), however, is that it requires quite large sample sizes (\textgreater{}50), below which it loses statistical power.

Two \(p\) values determine the statistical significance of \(D\), one each for the null hypotheses that \(D = 0\) (phylogenetic signal present) and \(D = 1\) (the character is distributed randomly relative to phylogenetic structure). These \(p\) values are obtained by comparing the observed \(D\) score to the two distributions of simulated \(D\) scores described above (\(\Sigma d_r\) and \(\Sigma d_b\)). The fraction of randomly simulated \(D\) scores smaller than observed \(D\) is taken as the \(p\) value for \(H_{0(D=1)}\). Likewise, the proportion of the simulated \(D\) scores greater than the observed \(D\) value is the \(p\) value for \(H_{0(D=0)}\).

\hypertarget{materials}{%
\section{Materials}\label{materials}}

Our study measures phylogenetic signal in a variety of types of phonotactic characters, extracted using semi-automated methods from wordlists within the Pama-Nyungan family (Australia). Throughout, we take the \emph{doculect} to be our unit of study. A doculect is a language variety as documented in a given resource (Cysouw \& Good \protect\hyperlink{ref-cysouw_towards_2007}{2007}; Good \& Cysouw \protect\hyperlink{ref-good_languoid_2013}{2013}). That is to say, we treat each wordlist as its own unit of study, without making any claims about the status of the documented language variety's status as a language or dialect. This agnosticism is advantageous in phylogenetic studies, since the terms `language' and `dialect' imply something about the relationship of a documented language variety to other documented language varieties, and a commitment to one term or the other therefore represents a phylogenetic assumption.

\hypertarget{sample}{%
\subsection{Language sample}\label{sample}}

Pama-Nyungan is by far the largest language family on the Australian continent, covering nearly 90\% of its landmass (everywhere except for three areas: part of the Top End, part of the Kimberley, and the whole of Tasmania) and encompassing around two-thirds of the languages present at the time of European settlement (Bowern \& Atkinson \protect\hyperlink{ref-bowern_computational_2012}{2012}: 817). Pama-Nyungan was first proposed and named by Kenneth Hale (Wurm \protect\hyperlink{ref-wurm_aboriginal_1963}{1963}: 136) and it has been the subject of considerable historical linguistic study since this time. Although the family has presented some challenges for historical linguistics, the phylogenetic unity of Pama-Nyungan has been established on traditional historical linguistic grounds (Alpher \protect\hyperlink{ref-alpher_pama-nyungan:_2004}{2004}) with many subgroups identified within (for example, O'Grady, Voegelin \& Voegelin \protect\hyperlink{ref-ogrady_languages_1966}{1966}; Wurm \protect\hyperlink{ref-wurm_languages_1972}{1972}; Austin \protect\hyperlink{ref-austin_proto-kanyara_1981}{1981}). For an overview of the history of Pama-Nyungan classification, see Bowern \& Koch (\protect\hyperlink{ref-bowern_australian_2004}{2004} ch.~1--5) and Koch (\protect\hyperlink{ref-koch_historical_2014}{2014}). Bowern \& Atkinson (\protect\hyperlink{ref-bowern_computational_2012}{2012}) perform a computational phylogenetic analysis of Pama-Nyungan using lexical data from 194 language varieties, providing for the first time a fully bifurcating phylogeny of the entire Pama-Nyungan family. Bouckaert, Bowern \& Atkinson (\protect\hyperlink{ref-bouckaert_origin_2018}{2018}) subsequently perform a phylogeographic analysis using the same dataset, but refined and expanded to 306 language varieties and including a geographic element to estimate the point of origin and spread pattern of the family through time and space.

The Pama-Nyungan family provides an excellent test case for this study. It holds practical advantages which make the task of phonological comparison easier, but it also provides us with a deliberately high bar to clear from a theoretical perspective. Both of these features are a result of the unusual degree of phonological homogeneity observed among Australian languages. Australian languages have been noted for a degree of similarity between phonological inventories of contrastive segments that is exceptional and unexpected in light of the phylogenetic and geographical breadth of the family, the level of diversity observed in vocabulary and aspects of grammar, and the level of phonological diversity found in comparably-sized families of languages elsewhere in the world. This has been noted as early as Schmidt (\protect\hyperlink{ref-schmidt_gliederung_1919}{1919}) and in more recent times by Capell (\protect\hyperlink{ref-capell_new_1956}{1956}), Voegelin et al. (\protect\hyperlink{ref-voegelin_obtaining_1963}{1963}), Dixon (\protect\hyperlink{ref-dixon_languages_1980}{1980}), Busby (\protect\hyperlink{ref-busby_distribution_1982}{1982}), Hamilton (\protect\hyperlink{ref-hamilton_phonetic_1996}{1996}), Baker (\protect\hyperlink{ref-baker_word_2014}{2014}), Bowern (\protect\hyperlink{ref-bowern_standard_2017}{2017}) and Round (\protect\hyperlink{ref-round_segment_2021}{2021}\protect\hyperlink{ref-round_segment_2021}{a}), among others. This curious level of homogeneity extends to phonotactics too (Dixon \protect\hyperlink{ref-dixon_languages_1980}{1980}; Hamilton \protect\hyperlink{ref-hamilton_phonetic_1996}{1996}; Baker \protect\hyperlink{ref-baker_word_2014}{2014}; Round \protect\hyperlink{ref-round_phonotactics_2021}{2021}\protect\hyperlink{ref-round_phonotactics_2021}{b}).

On one hand, the abundance of similar phonological inventories makes the task of comparison between them easier, because it limits the problem of \emph{dataset sparsity}. Consider a character coding the frequency of some sequence of two segments \(xy\) in a language: This character can only be compared between languages that contain both \(x\) and \(y\) segments in their inventories. If a language lacks either segment in its inventory, then the character will be coded as absent or missing (as distinct from 0, where a language possesses both segments but never permits them in sequence). We expect fewer missing values in Australia, where languages tend to share a large proportion of directly comparable segments, when compared to other parts of the world where we would expect to see many more missing values.

On the other hand, an ostensibly high degree of phonological homogeneity, in spite of considerable phylogenetic diversity, presents challenges for historical linguistics. Baker (\protect\hyperlink{ref-baker_word_2014}{2014}: 141) and Alpher (\protect\hyperlink{ref-alpher_pama-nyungan:_2004}{2004}: 103) have both written on the difficulties for historical reconstruction in Australia because of this. Moreover, a phylogeny implies some degree of historical divergence, but in the case of Australian languages, there would appear to be little by way of phonological divergence, let alone divergences which are phylogenetically patterned. We therefore choose to study an Australian language family as a deliberately difficult test case, where we expect the bar be set high with respect to detecting phylogenetic signal.

Gasser \& Bowern (\protect\hyperlink{ref-gasser_revisiting_2014}{2014}) counter prevailing views on Australian phonological homogeneity. They find that common assumptions, of the kind discussed above and commonly found repeated in reference grammars, mask a degree of variation which is otherwise revealed by, firstly, extracting data on segmental inventories directly from wordlists and, secondly, considering segmental frequencies extracted from wordlists. This result motivates our current approach; here, we are also concerned with matters of frequency, extracted directly from language wordlists. However, we look at different kinds of characters, pertaining not to single segments but to biphones, and consider them with respect to their phylogenetic implications.

\hypertarget{wordlists}{%
\subsection{Wordlists}\label{wordlists}}

Our Pama-Nyungan phonotactic data is extracted from 111 wordlists which are part of a database under development by the last author (Round \protect\hyperlink{ref-round_ausphon-lexicon_2017}{2017}\protect\hyperlink{ref-round_ausphon-lexicon_2017}{b}), extending the Chirila resources for Australian languages (Bowern \protect\hyperlink{ref-bowern_chirila:_2016}{2016}). In this study we restrict our attention to the most accurate sources available, and use only lexical data that is compiled by trained linguists and for which the underlying dataset is available in published or archived form. Additionally, we restrict our sample to wordlists containing a minimum of 250 words. We include this cut-off since measurement accuracy is a concern for smaller wordlists. A documented wordlist is necessarily only a subset of the complete lexicon of a language and it is unclear how big a wordlist must be before frequency statistics begin to stabilize around a sufficient level of accuracy. There is some work in this space concerning frequences of single segments (Dockum \& Bowern \protect\hyperlink{ref-dockum_swadesh_2019}{2019}), suggesting a rapid decline in the accuracy of phoneme frequencies as wordlists drop below 250 words. Longer wordlists will always be better, however we select 250 words as a reasonable compromise which maintains a generally broad coverage of Pama-Nyungan languages. We return to the subject of wordlist sizes and potential implications for our results in Section \ref{overall-robustness}

\begin{figure}

{\centering \includegraphics[width=1\linewidth]{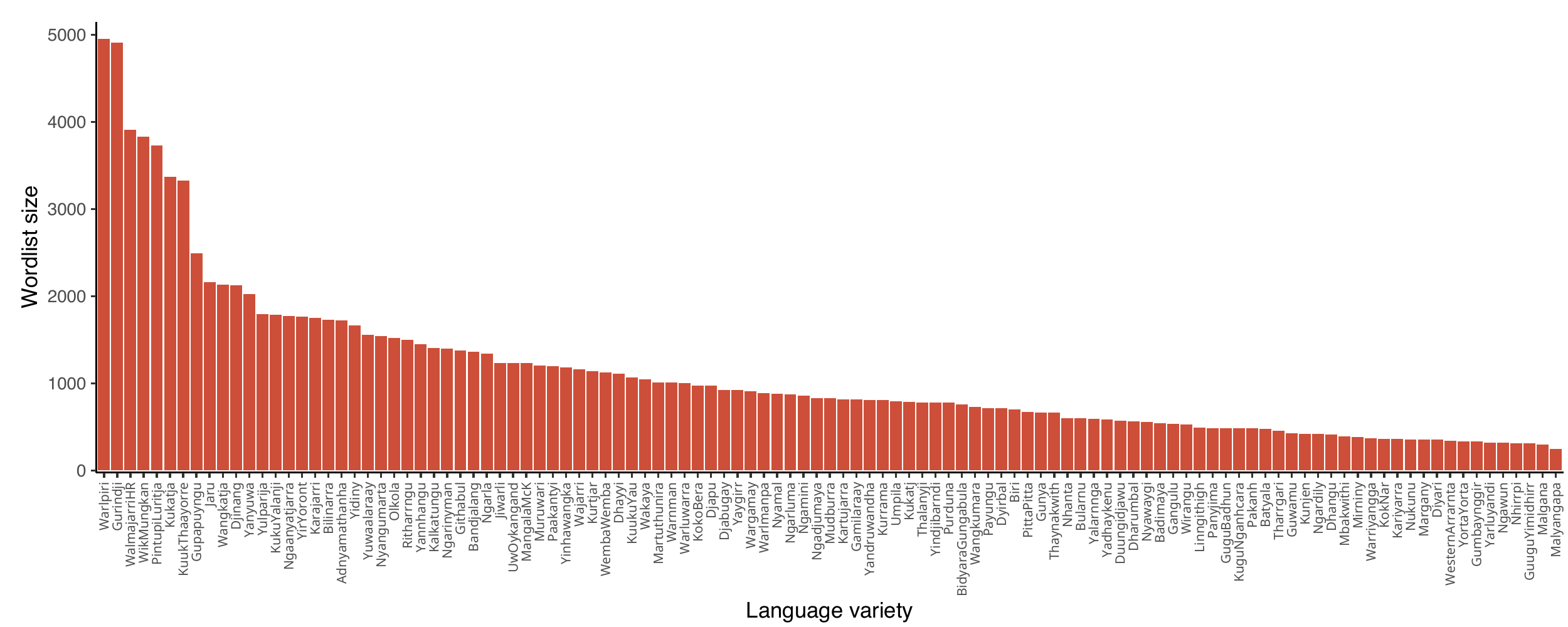} 

}

\caption{Lexicon sizes.}\label{fig:lex-size}
\end{figure}

Bibliographic details for all underlying data is available in Section S2 of the Supplementary Information. Owing to differences in the length of primary sources, there is considerable diversity in the size of the lexicons we use. As shown in Figure \ref{fig:lex-size}, the difference from smallest to largest is over an order of magnitude (min. 250, max. 4955), with the middle fifty percent between 509 and 1367 items. Mean lexicon size is 1112 (\(SD\) 916))

Original source data, which is typically orthographic and, if digital, is sometimes mixed with metadata or other extraneous material, has undergone extensive data scrubbing, conversion to phonemic form using language-specific orthography profiles (Moran \& Cysouw \protect\hyperlink{ref-moran_unicode_2018}{2018}), and additional automated and manual error checking. These procedures ensure basic data cleanliness. Separately however, it has long been recognized that the segmental-phonological analysis of languages is a non-deterministic process (Chao \protect\hyperlink{ref-chao_non-uniqueness_1934}{1934}; Hockett \protect\hyperlink{ref-hockett_problem_1963}{1963}; Hyman \protect\hyperlink{ref-hyman_universals_2008}{2008}; Dresher \protect\hyperlink{ref-dresher_contrastive_2009}{2009}). Two linguists faced with the same data may produce different analyses, not due to error but due to different applications of the very many analytic criteria that figure into any analysis of segments. Consequently, the cross-linguistic phonological record varies not only according to language facts per se, but due also to variation in the practice of linguistic analysis. Recent literature (Lass \protect\hyperlink{ref-lass_vowel_1984}{1984}; Hyman \protect\hyperlink{ref-hyman_universals_2008}{2008}; Van der Hulst \protect\hyperlink{ref-van_der_hulst_phonological_2017}{2017}; Round \protect\hyperlink{ref-round_matthew_2017}{2017}\protect\hyperlink{ref-round_matthew_2017}{a}; Kiparsky \protect\hyperlink{ref-kiparsky_formal_2018}{2018}) emphasizes the value of normalizing source descriptions prior to the analysis of cross-linguistic phonological datasets. This is not an information destroying process---it does not `standardize' languages---but it may shift information from one part of the representation (e.g., contrast between individual symbols) to another (e.g., contrasts between sequences of symbols), in order that information is located in a comparable way across the languages in the dataset, and therefore is more amenable to comparative analysis. Our wordlist data is normalized is this sense. Complex segments are split into simple sequences (e.g., prenasalized stops are split into a homorganic nasal + stop sequence); long vowels are represented as a sequence of identical short vowels, and vowel-glide-vowel sequences in which the glide is homorganic with either vowel are normalized to vowel-vowel; fortis consonants are represented as a sequence of identical short consonants, and positionally neutralized fortis/lenis stops as singletons; laminal consonants which do not figure in a pre-palatal versus dental opposition are represented as palatal, and rhotic glides which do not figure in an alveolar versus post-alveolar opposition are represented as post-alveolar (see also Round \protect\hyperlink{ref-round_phonemic_2019}{2019}\protect\hyperlink{ref-round_phonemic_2019}{a}; Round \protect\hyperlink{ref-round_australian_2019}{2019}\protect\hyperlink{ref-round_australian_2019}{b}). The phonotactic character sets used in this study were extracted from these normalized, comparably segmented wordlists.

\hypertarget{ref-phylogeny}{%
\subsection{Reference phylogeny}\label{ref-phylogeny}}

The reference phylogeny we use is a maximum clade credibility tree\footnote{Bayesian phylogenetic methods return not a single phylogenetic tree but a posterior distribution of many possible trees. These trees can be summarized into a single maximum clade credibility tree with confidence levels for each node in the tree, pertaining to how frequently that node appears in the posterior sample. It is the maximum clade credibility tree that we use for a reference phylogeny in this study. See Bowern \& Atkinson (\protect\hyperlink{ref-bowern_computational_2012}{2012}) and Bouckaert, Bowern \& Atkinson (\protect\hyperlink{ref-bouckaert_origin_2018}{2018}) for a full explanation of the methods used to infer the phylogenies considered in this section.} of 285 Pama-Nyungan language varieties inferred using lexical cognate characters by the second author (Figure S1, Supplementary Information). It was inferred independently of this study, prior to this study's conception and without the involvement of the first and third authors. It was inferred using the same Stochastic Dollo model as Bowern \& Atkinson (\protect\hyperlink{ref-bowern_computational_2012}{2012}), but with an expanded and refined dataset. Further details of the model and phylogeny construction are described in Bowern \& Atkinson (\protect\hyperlink{ref-bowern_computational_2012}{2012}), Bowern (\protect\hyperlink{ref-bowern_pama-nyungan_2015}{2015}) and Bouckaert, Bowern \& Atkinson (\protect\hyperlink{ref-bouckaert_origin_2018}{2018}). The cognate data used to infer the reference phylogeny is available on Zenodo (Bowern \protect\hyperlink{ref-bowern_pama-nyungan_2018}{2018}\protect\hyperlink{ref-bowern_pama-nyungan_2018}{b}). See Section S1 of the Supplementary Information for more information on the reference phylogeny.

We considered a reference tree from a newer phylogeographic analysis of Pama-Nyungan based on largely the same data plus further expansion to 304 doculects and continued refinement (Bouckaert, Bowern \& Atkinson \protect\hyperlink{ref-bouckaert_origin_2018}{2018}), however, we opted against its use for this particular study. The reason for this is that, although Bayesian inference of phylogenetic tree topology is considered generally robust to the levels of lexical borrowing observed among Pama-Nyungan languages (Greenhill, Currie \& Gray \protect\hyperlink{ref-greenhill_does_2009}{2009}; Bowern et al. \protect\hyperlink{ref-bowern_does_2011}{2011}), borrowing still has the effect of reducing branch lengths across the tree (Greenhill, Currie \& Gray \protect\hyperlink{ref-greenhill_does_2009}{2009}). This effect, and consequently the accuracy of branch length estimates, is equally applicable to both trees considered here. However, the geographic element in the phylogeographic study uses, in part, branch lengths to model geographic dispersal. The posterior distribution of trees, which is jointly informed by cognate data and geography, may therefore show a bias towards geographically proximal languages whose apparent divergence times have been reduced by high rates of borrowing. Thus, although branch length estimates will be impacted by borrowing in any phylogenetic study of Pama-Nyungan, there is more chance of borrowing affecting topology in the phylogeographic study.

We consider it unlikely that the overall conclusions of the study would be altered by the choice of which version of the Pama-Nyungan phylogeny we use as a reference tree. Each of the studies referenced above produced highly congruent Pama-Nyungan phylogenies (see Bouckaert, Bowern \& Atkinson \protect\hyperlink{ref-bouckaert_origin_2018}{2018} for a detailed comparison). Furthermore, Bouckaert, Bowern \& Atkinson (\protect\hyperlink{ref-bouckaert_origin_2018}{2018}) features fixed clade priors based on subgroups identified in earlier studies, so topological differences are constrained to some extent by design. Nevertheless, the accuracy of the reference tree is a key assumption of the methods we use in this study, and thus phylogenetic uncertainty is an important consideration. We return to this point in Section \ref{overall-robustness} and evaluate the overall robustness of our results to phylogenetic uncertainty by replicating a subset of phylogenetic signal tests over a posterior sample of trees.

As discussed above, we treat each wordlist in our study as its own doculect. The reference tree in this study was inferred using a similar approach, while remaining less-commital about the particular status of the unit of analysis. Resources were sometimes combined for a particular language, but they are also frequently broken up into separate units, particularly when the resources come from different authors and different time periods. We have taken care to match the wordlists in this study to their exact or best corresponding tip in the reference tree. In most cases, the wordlists we use here are the same as those used to generate the cognacy judgements used to infer the reference phylogeny. In other cases, we use a different source to the one used for the reference phylogeny but there is, nevertheless, a straightforward one-to-one mapping between the language variety our wordlist represents and a corresponding tip in the tree. In one case, our Mudburra source (Nash et al. \protect\hyperlink{ref-nash_mudburra_1988}{1988}) matches neither of the sources for the two Mudburra tips in the reference phylogeny. However, the two varieties in the reference phylogeny have the same date. This entails that when either of them is removed from the tree, the exact same result is obtained in terms of tree geometry, which is what is significant for our investigation. Accordingly, we remove one and match our source to the other.

\hypertarget{phy-sig-bin}{%
\section{Phylogenetic signal in binary phonotactic data}\label{phy-sig-bin}}

In the simplest case, the phonotactics of different languages may be compared in terms of which sequences of two segments (\emph{biphones}) they permit and which they do not. If claims about the relative homogeneity of phonotactic constraints in Australian languages holds, then we would expect this kind of comparison to yield little, if any, phylogenetic signal.

In this test, we construct the dataset as follows: We automatically extract from all wordlists every unique sequence of two segments---or more accurately, sequences of \(xy\) where each of \(x\) and \(y\) is either a phonological segment or a word boundary `\#'. Each sequence becomes a character (variable) in the dataset, for which every language receives a binary value: 1 if the sequence \(xy\) is found in the language's wordlist (even if only once); 0 if the language contains both segments in its inventory but the sequence \(xy\) never appears in its wordlist; or, NA (not applicable, missing) if the language does not contain one (or both) of either segment \(x\) or \(y\) in its inventory (and therefore \emph{a prioi} cannot contain the sequence \(xy\)). Binary data of this kind represents sequence permissibility: Where a language contains both segments in its inventory, it will either permit them to appear together in sequence or it will not. In this respect, the information encoded by these characters is similar to what one might find in the phonotactics section of a descriptive grammar, where one often encounters a description in prose and/or a basic tabulation of which segments are permitted and where, within syllable and word structures. However, this kind of information is also, in a sense, quite coarse-grained, since there are only two possible values. A sequence which is very common in one language will be coded in exactly the same way as a sequence which only appears a handful of times in another language.

We apply the \(D\) test individually to each character in the dataset that meets two conditions: at least 50 non-missing values (due to the aforementioned reliability issue with sample sizes smaller than this) and at least one instance of variation (we do not test characters where all languages share identical 1 or 0 values). Given the extensive history of description of Australian languages as phonotactically homogenous, our prior expectation is that testing binary data will fail to yield significant phylogenetic signal. Indeed, we might expect that \(D\) will fall significantly below 0, indicating that values are clumped among tips on the reference phylogeny even more conservatively than would be expected if they had evolved in the same phylogenetic pattern as lexical data.

To evaluate the statistical significance of \(D\) for any given character, a \(p\) value is estimated for each of two null hypotheses: The null hypothesis that \(D = 1\) (\(H_{0(D=1)}\), character values are distributed randomly with regards to phylogeny) and the null hypothesis that \(D = 0\) (\(H_{0(D=0)}\), character values are distributed as could be expected if the character has evolved along the phylogeny according to a Brownian motion model). Each \(p\) value is calculated using a randomization procedure: A random distribution of \(D\) scores is acquired by randomly shuffling character values among tips on the tree for 10,000 permutations. The conventional cutoff for statistical significance is \(p = 0.05\). Here, we use the corresponding Bonferroni-corrected cutoff of 0.025.\footnote{Bonferroni correction is used because the conventional threshold for statistical significance, 0.05, represents the expected chance of a false discovery (false positive). This figure is known as the type I error rate (\(\alpha\)). The chance of a false discovery is multiplied when multiple tests are carried out. Bonferroni correction, which involves dividing the threshold for statistical significance by the number of tests being conducted, ensures that the chance of observing a false positive in any of the set of tests remains at the conventional rate, \(\alpha = 0.05\). In our case, two null hypotheses are tested for each character, hence we divide the threshold for statistical significance by two, ensuring the chance of a false positive for any particular character is 0.05.} For any given character, there are six possible results:

\begin{itemize}
\tightlist
\item
  \(D\) is significantly below 0. Character values are even more tightly clumped among close relatives than Brownian motion alone would lead us to expect.
\item
  \(D\) is significantly below 1 and not significantly different from 0. The data patterns phylogenetically, i.e., there is phylogenetic signal.
\item
  \(D\) is significantly above 0 and below 1. The data is neither clearly random nor clearly phylogenetic.
\item
  \(D\) is significantly above 0 and not significantly different from 1. It is consistent with randomness, not phylogeny.
\item
  \(D\) is significantly above 1. It is even more dispersed than expected via a random process.
\item
  \(D\) is not significantly distinct from 0 nor 1. The patterning of the data is indeterminate, and cannot be distinguished from randomness nor from Brownian phylogenetic evolution.
\end{itemize}

To summarize, the testing procedure proceeds as follows. For each binary biphone character, if the character has at least 50 non-NA values \emph{and} the character has at least one `1' and one `0' value (i.e.~not every value is identical), then test as follows (otherwise discard):

\begin{itemize}
\tightlist
\item
  Calculate \(D\)
\item
  Conduct randomisation procedure to calculate \(p\) for \(H_0: D = 0\)
\item
  Conduct randomisation procedure to calculate \(p\) for \(H_0: D = 1\)
\end{itemize}

A result is interpreted from the combination of \(D\) and two \(p\) values.

\hypertarget{phy-sig-bin-results}{%
\subsection{Results for binary phonotactic data}\label{phy-sig-bin-results}}

\begin{figure}

{\centering \includegraphics[width=0.66\linewidth]{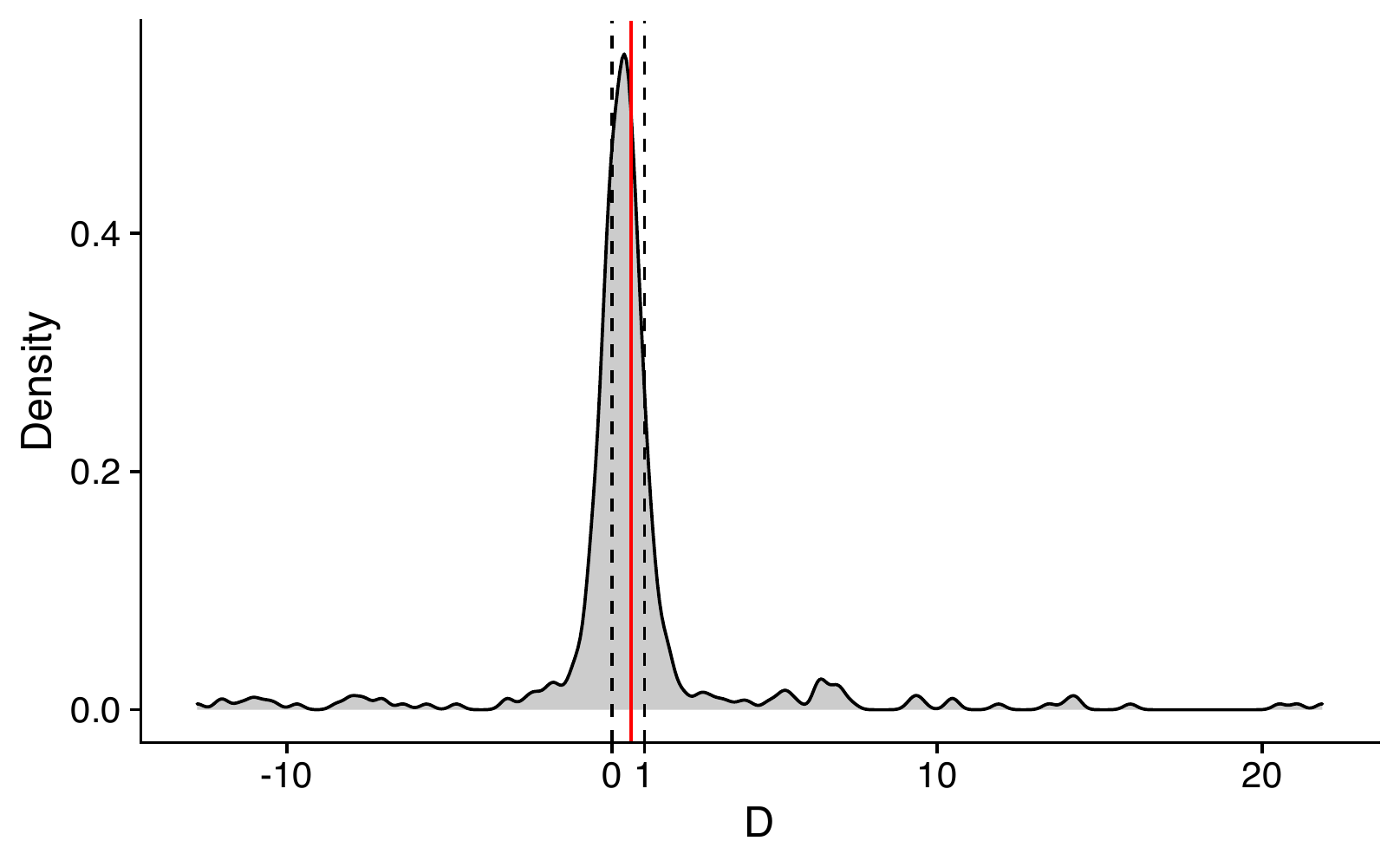} 

}

\caption{Density of $D$ estimates for binary biphone characters. Dotted lines mark $D=0$, the phylogenetic expectation, and $D=1$, the random expectation. Mean $D$ for all characters is 0.43, marked in red.}\label{fig:d-density}
\end{figure}

We estimate \(D\) for 415 biphone characters using a script based on the \emph{phylo.d} function in the \emph{caper} package (Orme et al. \protect\hyperlink{ref-orme_caper:_2013}{2013}), implemented in the statistical software \emph{R} (R Core Team \protect\hyperlink{ref-r_core_team_r:_2017}{2017})\footnote{The dataset for this and subsequent tests are available on Zenodo at \url{http://doi.org/10.5281/zenodo.3610089}. This repository also includes a full table of results and the R scripts used to perform the analysis and produce figures for the paper. See Section S3 of the Supplementary Information for usage instructions and a full description of these materials.}. As described in Section \ref{phylo-sig}, \(D\) is calibrated by simulating character evolution under two models---one where the character evolves at random relative to phylogeny and one where the character evolves following a Brownian motion threshold model. In this study, we conduct 10,000 permutations of each model for each character. The 415 \(D\) values cluster centrally around a mean of 0.59. The distribution is leptokurtic (kurtosis = 10.9), meaning there are more outliers relative to a normal distribution, making the distribution appear as a tall, narrow peak with long tails (Figure \ref{fig:d-density}). The standard deviation is large (3.64).

\begin{figure}

{\centering \includegraphics[width=0.9\linewidth]{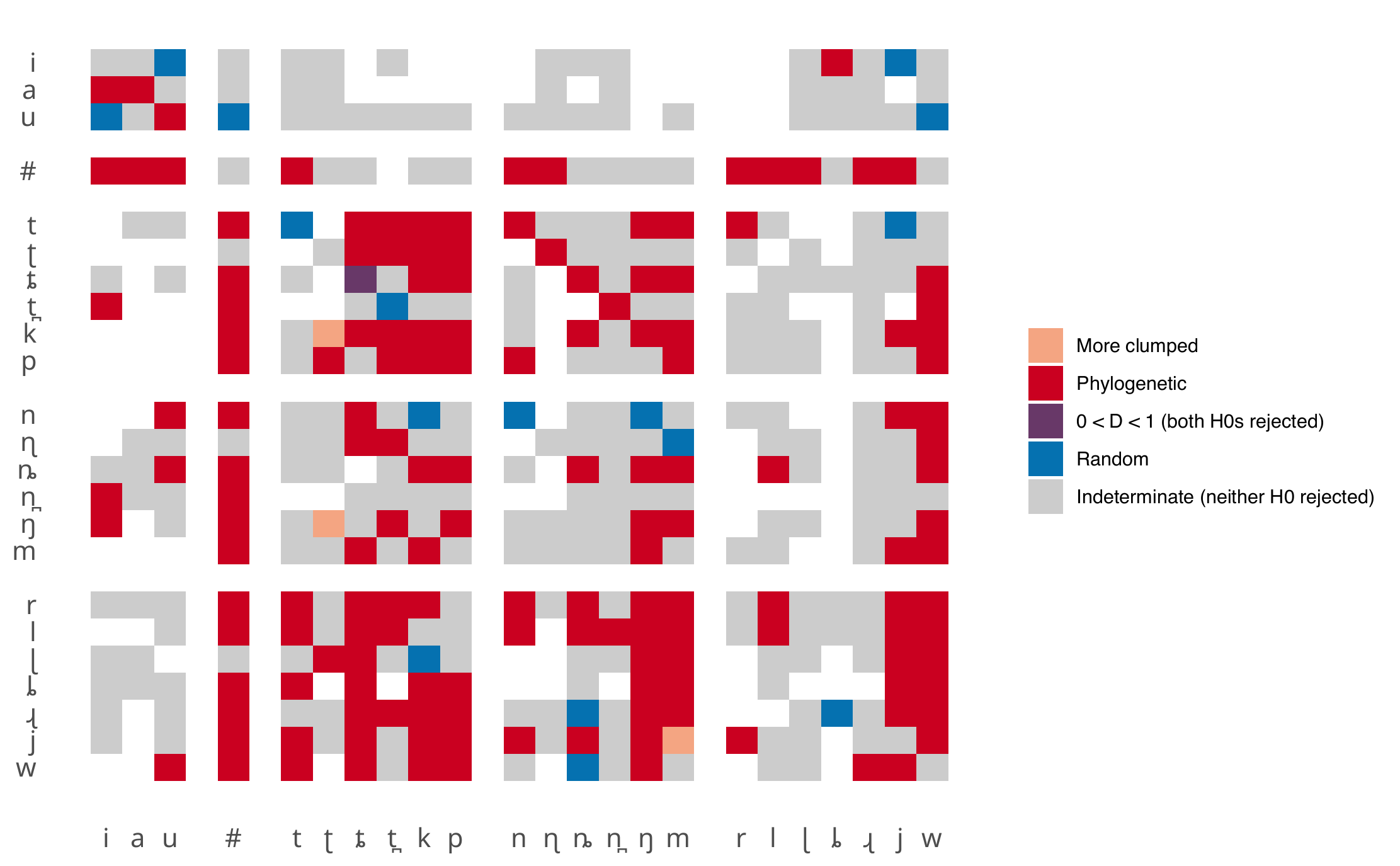} 

}

\caption{Phylogenetic signal significance testing for binary biphone characters. This grid colour-codes each biphone character according to the results of its respective significance tests. The grid is arranged such that the vertical axis represents the first segment of the biphone and the horizontal axis represents the second segment. Besides a tendency for phonotactic restrictions at word boundaries to show phylogenetic signal, few patterns stand out.}\label{fig:d-swatch}
\end{figure}

The \(D\) test was of indeterminate significance for half of all characters (238 characters, 57\% of the dataset). The \(D\) scores for 157 characters (38\% of the total dataset) show evidence of phylogenetic signal. Just 16 characters (4\%) show the opposite result, where the character is consistent with randomness and there is no phylogenetic signal present. Both null hypotheses are rejected for the remaining 4 characters. Of these, 3 are more clumped than their phylogenetic expectation and 1 falls somewhere between phylogenetic and random expectations (\(0 < D < 1\)). None are more dispersed than the random expectation. The distribution of these results among different biphone characters is plotted in Figure \ref{fig:d-swatch}. Note that we expect around 5\% of null hypothesis rejections (approximately 18 of 180 rejected null hypotheses) to be false discoveries. Nevertheless, when considering the whole dataset as an ensemble rather than each character individually, a general result can be discerned. The clearest conclusion is that binary, permissibility-based characters tend to be low yielding in information, giving a statistically significant outcome in fewer than half of cases. Nevertheless, where a significant result can be determined, phylogenetic signal does tend to be present---to a degree that is perhaps surprising in light of previous literature describing the relative homogeneity of Australian phonotactic restrictions and their lack of utility in historical endeavours. This result suggests there may be a greater degree of historical information contained in Pama-Nyungan phonotactics than previously thought. However, it may be that a finer-grained approach to data extraction is needed in order to detect it.

These results can be compared to two earlier studies performing the same test on much smaller samples of languages. In the first, Macklin-Cordes \& Round (\protect\hyperlink{ref-macklin-cordes_high-definition_2015}{2015}) find no evidence for phylogenetic signal in the Yolngu subgroup of Pama-Nyungan---rather, data are significantly over-clumped, suggesting a higher degree of conservatism in phonotactic restrictions relative to lexical data. They fail to reject a null hypothesis of \(D = 0\) for Ngumpin-Yapa, suggesting there may be a degree of phylogenetic signal in the Ngumpin-Yapa dataset. However, the pilot study results should be treated with caution---particularly the failure to reject the \(D = 0\) null hypothesis in the case of Ngumpin-Yapa---due to the small sample sizes (10 languages for Ngumpin-Yapa, 7 for Yolngu), well below the minimum of 50 taxa recommended by Fritz \& Purvis (\protect\hyperlink{ref-fritz_selectivity_2010}{2010}). Dockum (\protect\hyperlink{ref-dockum_phylogeny_2018}{2018}) performs the same analysis using biphone characters from 20 Tai lects of the Kra-Dai family. In contrast to Macklin-Cordes \& Round (\protect\hyperlink{ref-macklin-cordes_high-definition_2015}{2015}), Dockum finds some evidence of phylogenetic signal in the Tai data and suggests perhaps the earlier result was due to insufficient variation in that particular language sample rather than a limitation of binary biphone characters per se. Although the low information yield from binary data is to be expected, our results here appear to support Dockum's conclusion.

\hypertarget{phy-sig-bin-robustness}{%
\subsection{Robustness checks}\label{phy-sig-bin-robustness}}

\begin{figure}
\includegraphics[width=1\linewidth]{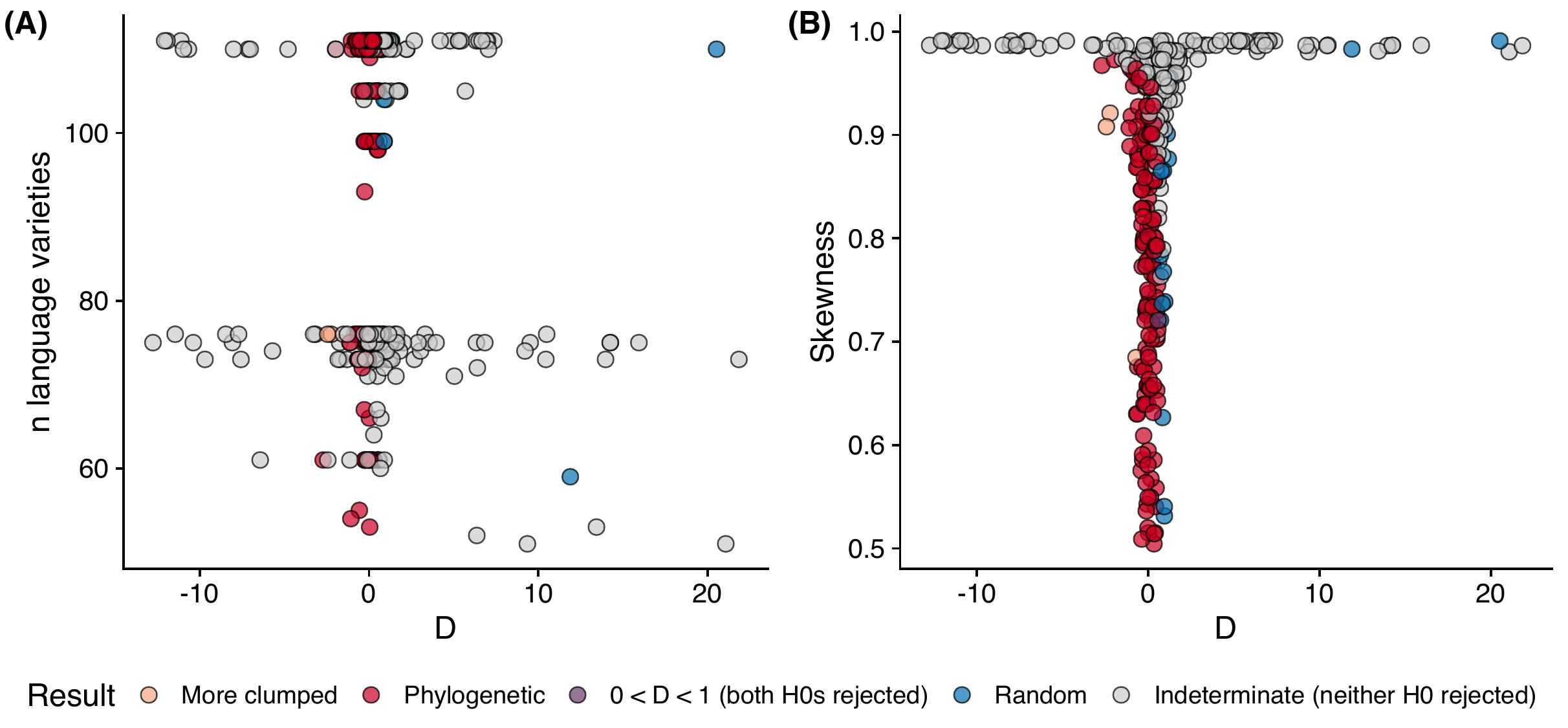} \caption{Scatterplot of $D$ scores against (A) the number of doculects with non-missing values for each character and (B) the skewing of the distribution of 1s and 0s for each character. $D$ clusters evenly around the mean regardless of the number of missing values. Variation in $D$ accelerates greatly among characters where all but 1 or a few doculects share the same value, but the results are overwhelmingly not significant.}\label{fig:d-scatterplots}
\end{figure}

\begin{figure}

{\centering \includegraphics[width=0.66\linewidth]{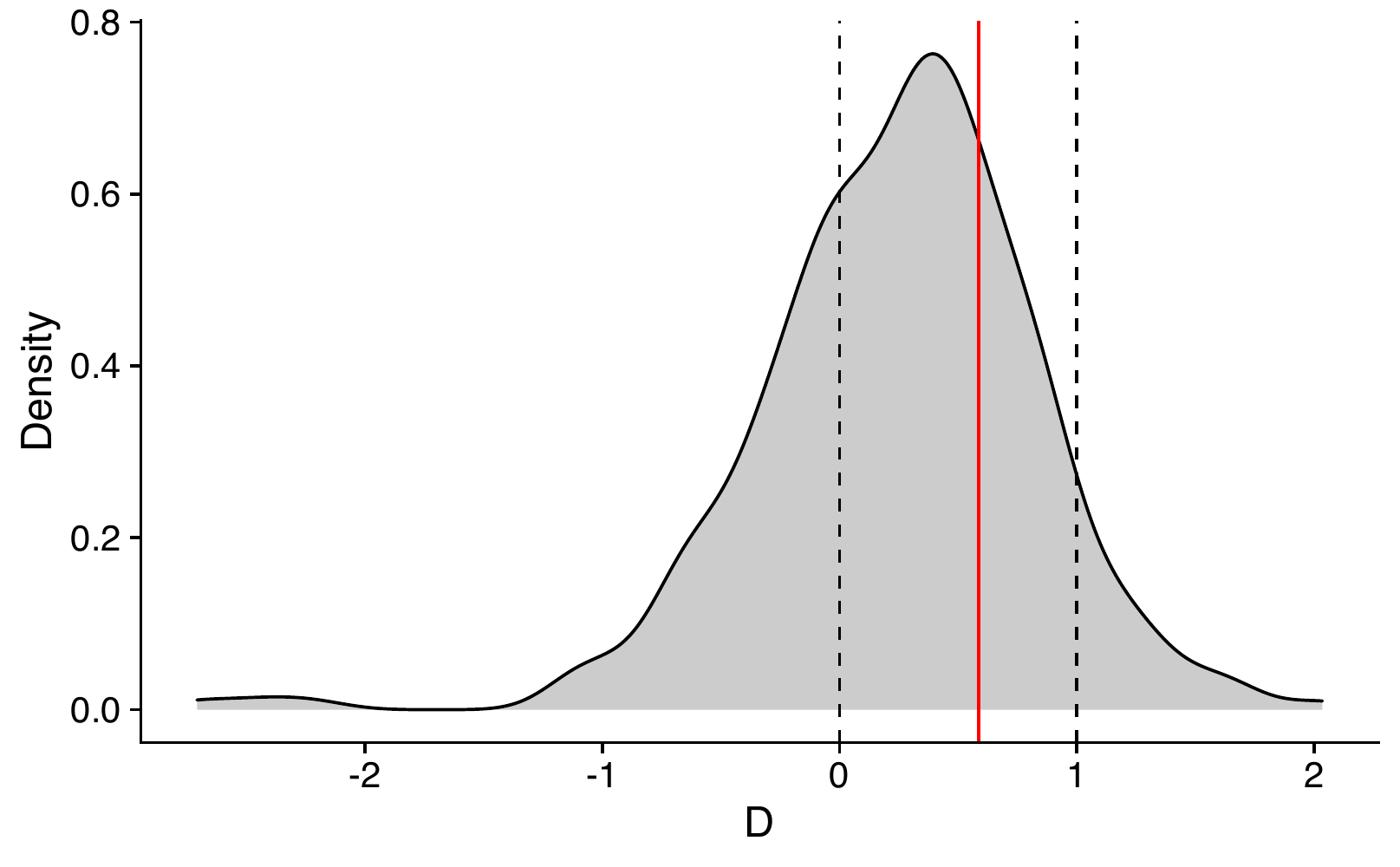} 

}

\caption{Density of $D$ estimates for binary biphone characters where character values are skewed less than 97-to-3.}\label{fig:d-density-filtered}
\end{figure}

\begin{figure}

{\centering \includegraphics[width=0.75\linewidth]{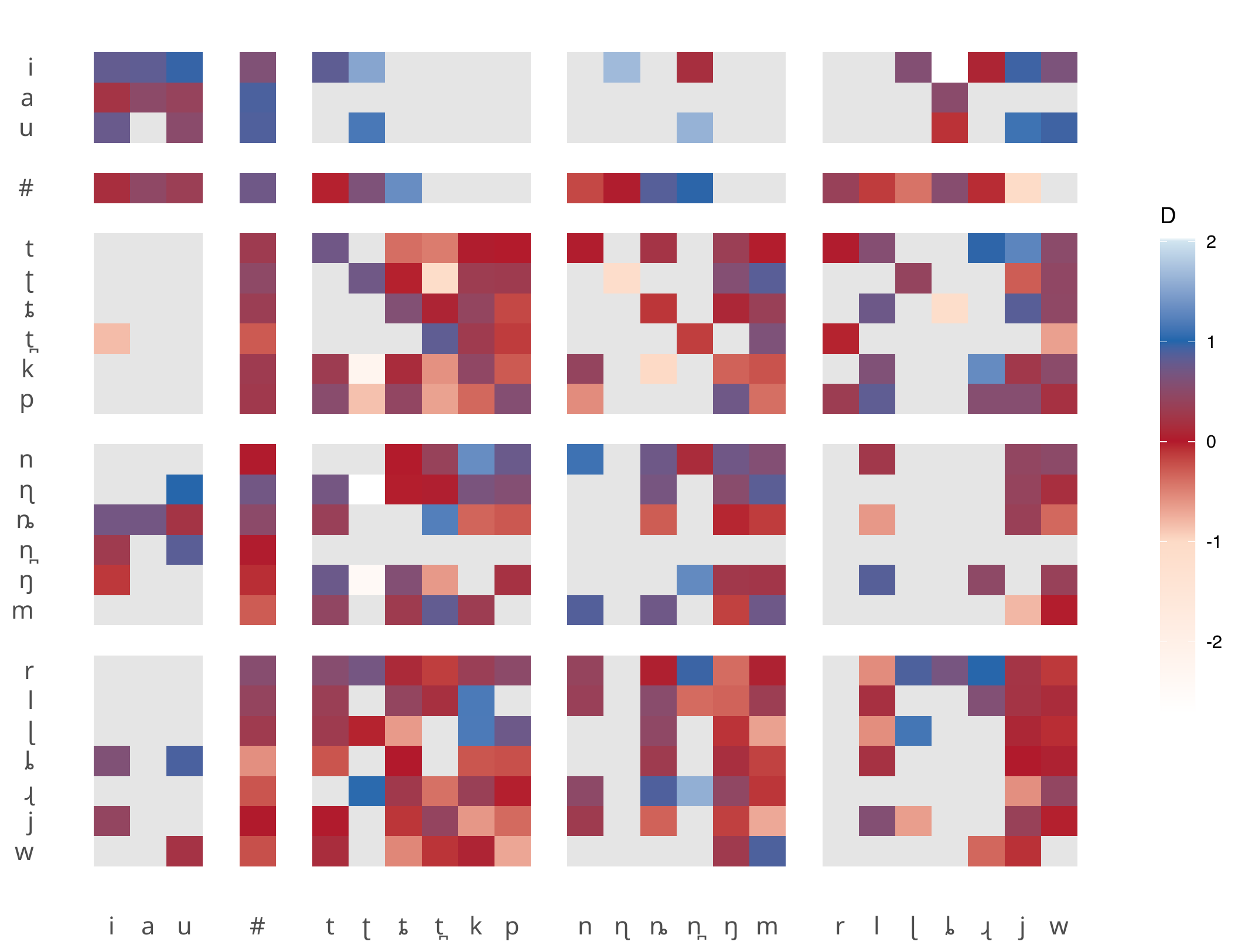} 

}

\caption{$D$ scores for binary biphone characters where character values are skewed less than 97-to-3. Red shades show a character's proximity to $D=0$---darker red indicates stronger phylogenetic signal. Blue shades show a character's proximity to $D=1$, where character values are distributed randomly. This heat grid shows the proximity of each individual character's $D$ value to $D=0$---its expectation if the character evolved along the phylogeny following a Brownian motion process (the vertical axis represents the first segment of the biphone, the horizontal axis represents the second segment).}\label{fig:d-swatch-filtered}
\end{figure}

Given a sufficient number of taxa for which data are available (\textgreater{}50), \(D\) scores should reflect a degree of phylogenetic signal present in the data, independently of tree size (the number of taxa) and shape (branching patterns). To check this, in Figure \ref{fig:d-scatterplots}(A) for each character we plot its \(D\) score against the number of doculects for which it had non-missing values. Irrespective of the number of doculects that supply non-missing values, the \(D\) scores appear to cluster centrally around mean \(D\) for the dataset, suggesting that \(D\) is not being unduly affected by missing values for particular characters. A second check leads to rather different results, however. We check whether skewed distributions of character values affects \(D\) scores (Figure \ref{fig:d-scatterplots}(B)). Here, we consider the distribution of 1s and 0s for each character and plot \(D\) against how skewed the distribution of character values is towards a particular value. For example, a character where there are 107 `1' values in the dataset and only 4 `0' values will have a skewing rating of 0.96 (the count of `1' values, 107, divided by 111 total observations). Here, we find that when the ratio of 1s and 0s for a character is highly unequal, estimates of \(D\) tend towards extreme magnitudes while also being unrevealing, that is, statistically distinguishable neither from 0 nor 1.\footnote{Note that it is a desirable feature of the \(D\) test that it should return a lack of significance when there is a near-complete lack of variability in the data for the test to evaluate. This is an issue with the data, not the test.} As described above, Australian languages are known for homogeneity in phonological inventories and phonotactic restrictions, so consequently there are many characters with skewed distributions affecting the results. In Figures \ref{fig:d-density-filtered}--\ref{fig:d-swatch-filtered}, we plot a subset of the \(D\) test results, restricted to characters with less skewing.

\hypertarget{phy-sig-cont}{%
\section{Phylogenetic signal in continuous phonotactic data}\label{phy-sig-cont}}

We test whether a higher degree of phylogenetic signal is detectable in continuous-valued biphone characters. As in the previous test, we take every possible sequence of two segments, or biphones, in our sample of 111 Pama-Nyungan wordlists. In this case, however, rather than simply coding for the presence of a biphone in a language's lexicon, we consider the relative frequencies of transitions between the segments in that biphone across the language's lexicon. For each biphone character, we take two values: The Markov chain forward transition probability---that is, for a biphone \(xy\), the probability of \(x\) being followed by \(y\), normalized over all instances of \(x\). This captures, if only in a basic way, our awareness that words do not consist of strings of independent segments, but rather the probability of observing some segment is very much dependent on what came before it. Secondly, we take Markov chain backward transition probabilities---that is, for the biphone \(xy\), the probability of \(y\) being preceeded by \(x\), normalized over all instances of \(y\) in the lexicon. The frequency characters we extract come from wordlists. This is advantageous in that they are somewhat independent of word frequency effects since each word is counted only once, in contrast to frequencies extracted from language corpora. On the other hand, speakers show sensitivity to phoneme frequencies in language use (for example, when coining novel words) (Coleman \& Pierrehumbert \protect\hyperlink{ref-coleman_stochastic_1997}{1997}; Zuraw \protect\hyperlink{ref-zuraw_patterned_2000}{2000}; Ernestus \& Baayen \protect\hyperlink{ref-ernestus_predicting_2003}{2003}; Albright \& Hayes \protect\hyperlink{ref-albright_rules_2003}{2003}; Eddington \protect\hyperlink{ref-eddington_spanish_2004}{2004}; Hayes \& Londe \protect\hyperlink{ref-hayes_stochastic_2006}{2006}; Gordon \protect\hyperlink{ref-gordon_phonological_2016}{2016}) so word frequency will likely have some effect on phoneme and biphone frequency even in a wordlist. Investigation of phylogenetic signal in frequency characters extracted from corpora versus wordlists may be a possibility for future study.

We quantify phylogenetic signal by estimating \(K\) (Blomberg, Garland \& Ives \protect\hyperlink{ref-blomberg_testing_2003}{2003}) individually for each character, using the \emph{multiPhylosignal} function, in the \emph{picante} package (Kembel et al. \protect\hyperlink{ref-kembel_picante:_2010}{2010}), in \emph{R} statistical software. The \(K\) test has somewhat greater statistical power than the \(D\) test, enabling us to apply the test to characters with as few as 20 non-missing values. Calculation of \(K\) works with non-zero values only, so zero values (where the language contains both segments \(x\) and \(y\) but \(x\) is never followed by \(y\), or vice versa) are considered not applicable and removed from calculation. A total of 490 characters (245 biphone forward transition probabilities and 245 backward transition probabilities) meet the criterion of at least 20 languages with non-missing and non-zero values for testing. Subsequently, to evaluate whether the level of phylogenetic signal is significant for a given character, we conduct Blomberg, Garland and Ives' (2003) randomization procedure with 10,000 random permutations per character.

To summarize, this testing procedure proceeds as follows. For each biphone frequency character, if the character has at least 20 non-NA values \emph{and} the character has at least two unique frequency values (i.e.~not every character value is the same), then test as follows (otherwise discard):

\begin{itemize}
\tightlist
\item
  Calculate \(K\)
\item
  Conduct randomisation procedure to calculate \(p\) for \(H_0: K = 0\)
\end{itemize}

\begin{figure}

{\centering \includegraphics[width=0.66\linewidth]{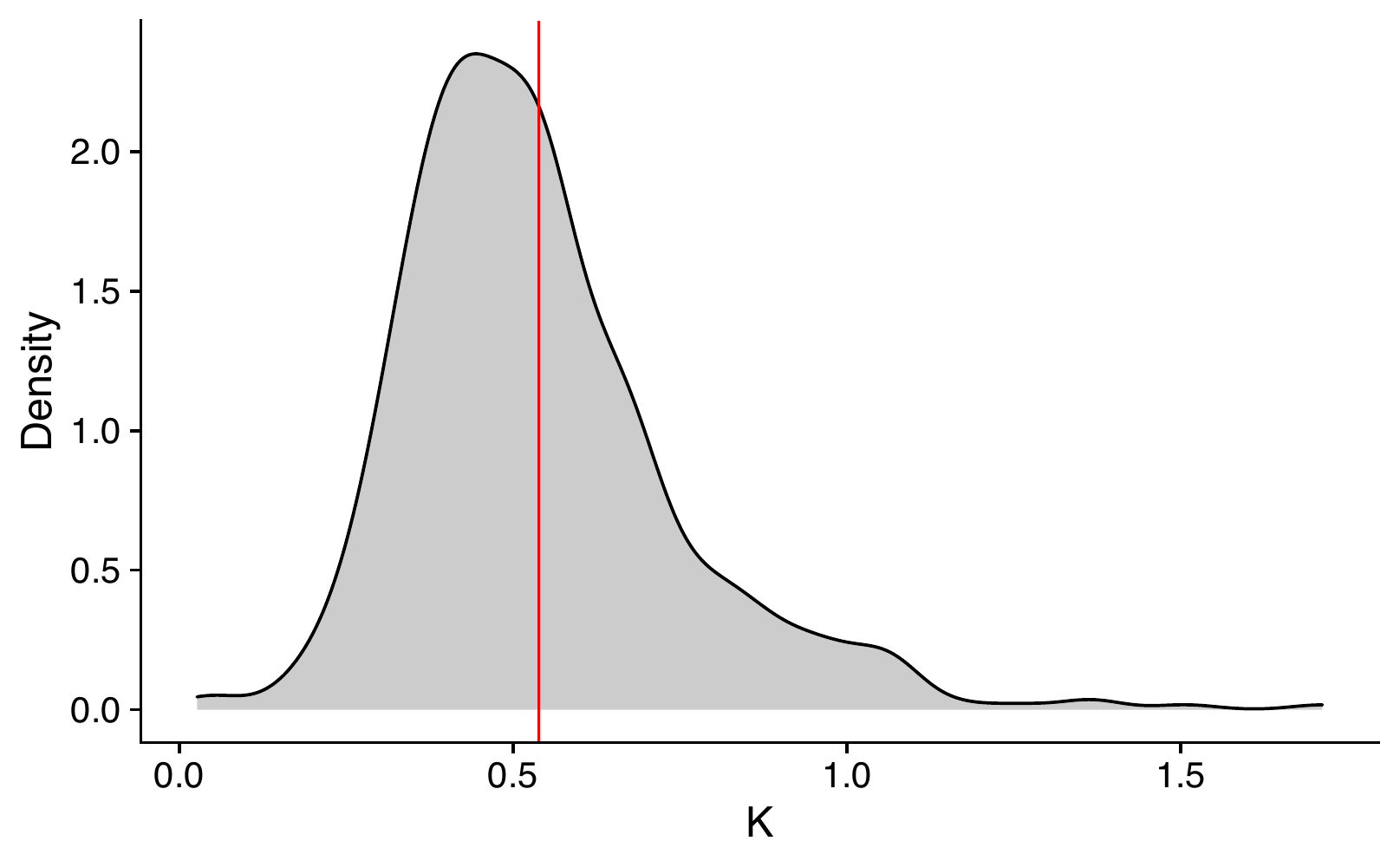} 

}

\caption{Density of $K$ scores for all frequency-based biphone character (frequencies of both forward and backward transitions between segments).}\label{fig:k-density}
\end{figure}

Mean \(K\) for all 490 characters is 0.54 (\(SD\) 0.21) (Figure \ref{fig:k-density}). This is comparable to certain physiological traits presented as examples of biological traits with a high degree of phylogenetic signal by Blomberg, Garland \& Ives (\protect\hyperlink{ref-blomberg_testing_2003}{2003}), for example, \(K=0.55\) for log body mass of primates. Using the Blomberg, Garland \& Ives (\protect\hyperlink{ref-blomberg_testing_2003}{2003}) randomisation procedure, we find a statistically significant degree of phylogenetic signal for 354 of 490 characters (180 forward transition characters, 174 backward transition characters), or 72\% of the total dataset.

\begin{figure}

{\centering \includegraphics[width=0.75\linewidth]{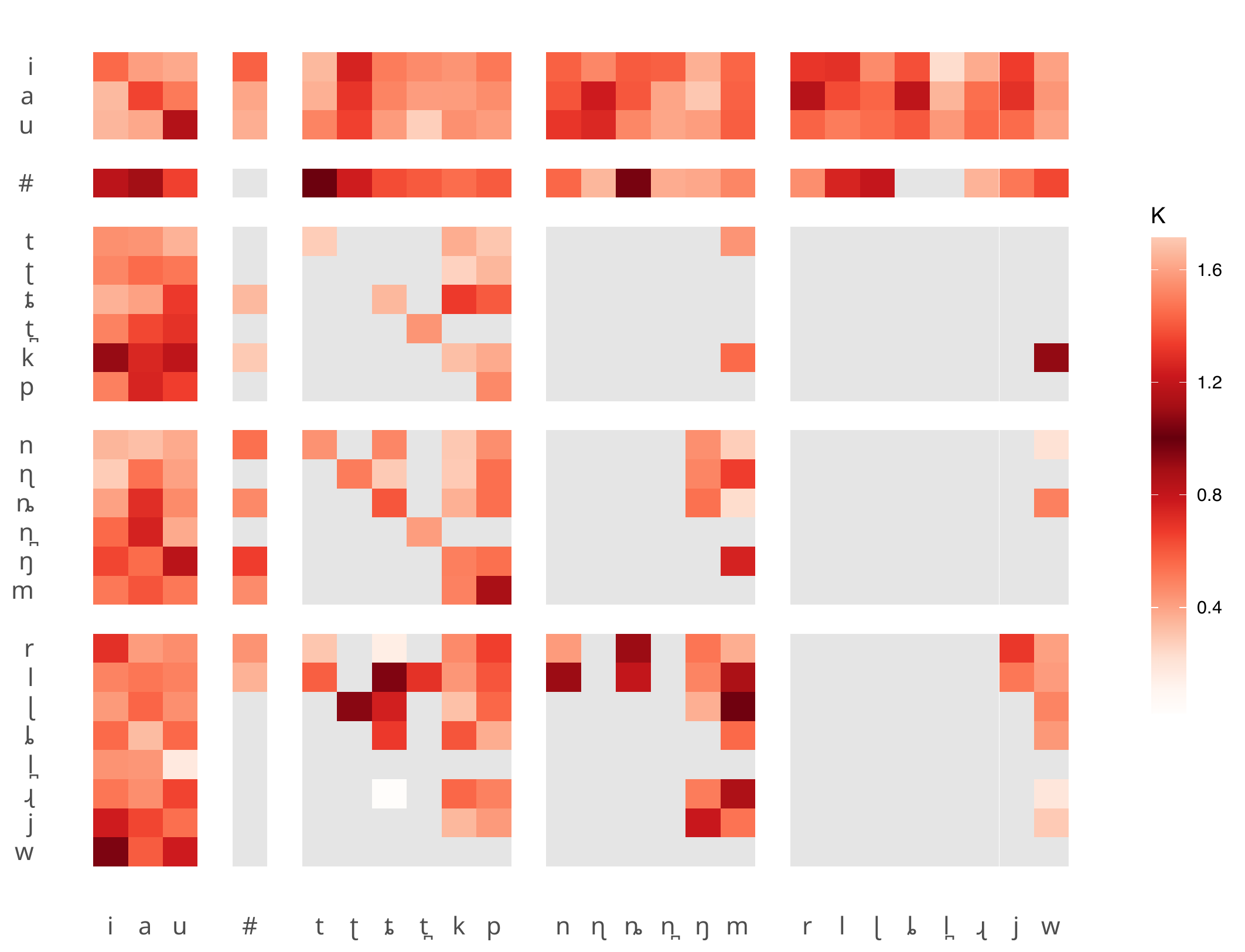} 

}

\caption{Phylogenetic signal for forward transition frequencies. This heat grid shows $K$ scores for biphone characters (forward transition frequencies only). Each square represents a biphone (where the first segment is listed on the vertical axis and second segment on the horizontal axis). Data points are taken from the frequencies of each biphone, $xy$, over the total frequency of segment $x$ in each language, and then phylogenetic signal $K$ is measured for each biphone. Darker red shades indicate a stronger degree of phylogenetic signa. As with the $D$ test, no clear pattern of high versus low $K$ scores stands out, although there is a high degree of phylogenetic signal in the dataset overall.}\label{fig:k-swatch}
\end{figure}

We consider whether phylogenetic signal is higher or lower in certain kinds of biphone characters. Figure \ref{fig:k-swatch} shows a matrix of \(K\) scores for forward transition characters, with rows and columns arranged by phonological natural class. No clear pattern stands out.

\hypertarget{phy-sig-cont-robustness}{%
\subsection{Robustness checks}\label{phy-sig-cont-robustness}}

Although \(K\) is intended to be a measure of phylogenetic signal that is independent of tree size and shape, tree size and shape can have some effect on results in practice (Münkemüller et al. \protect\hyperlink{ref-munkemuller_how_2012}{2012}). We wish to check that the Pama-Nyungan tree does not contain any unusual properties that could cause either the K statistic or the randomization procedure to perform unexpectedly. To do this, we allow simulated characters to evolve specifically along the Pama-Nyungan reference tree. We vary the model of evolution, between perfect Brownian motion along the entire tree and pure randomness generated directly at the tips of the tree, by mixing different strengths of Brownian phylogenetic signal and non-phylogenetic noise. 1000 traits are simulated at each percentage point interval for 100,000 total simulated traits (in other words, 1000 traits simulated with 100\% Brownian motion, then 1000 traits simulated with 99\% Brownian motion and 1\% randomness, and so on, until the traits evolve 100\% at random). Each simulated trait is then tested for statistical significance using the randomisation procedure described in Section \ref{phylo-sig}, with 1000 repetitions to determine a \(p\) value. In a robust testing scenario, \(K\) will scale appropriately between 0 and 1 according to the level of Brownian motion and random noise being simulated, and the randomisation procedure will distinguish between traits with and without a significant degree of phylogenetic signal with a satisfactory amount of Type I (false positive) and Type II (false negative) errors.

\begin{figure}

{\centering \includegraphics[width=0.9\linewidth]{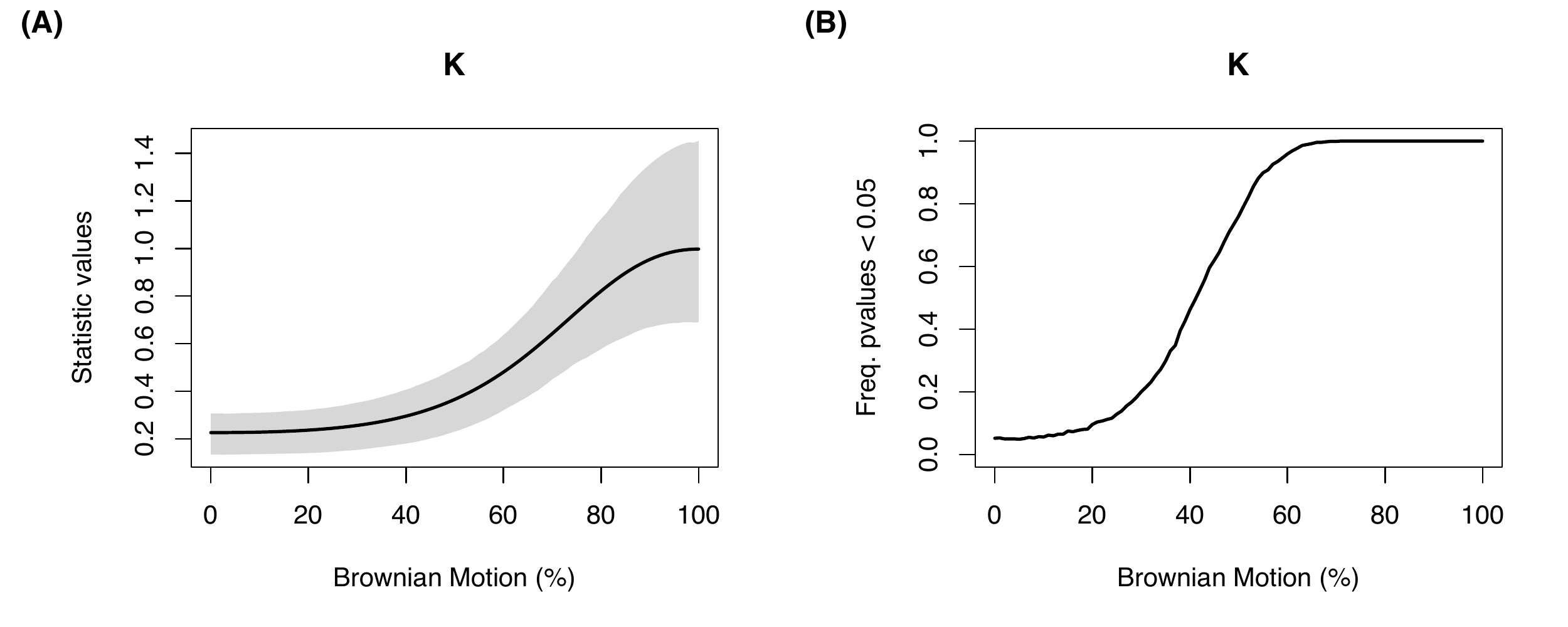} 

}

\caption{Behaviour of the $K$ statistic and randomization procedure with the Pama-Nyungan reference phylogeny. Artificial characters are simulated evolving along the phylogeny with varying levels of non-Brownian noise. Where a pure Brownian motion process operates, $K$ averages around 1, as expected. Where there is no Brownian process at all (and therefore no phylogenetic signal) $K$ is elevated to around 0.2---likely an artefact of this particular tree size and shape.}\label{fig:k-simulation-plots}
\end{figure}

The results are plotted in Figure \ref{fig:k-simulation-plots}. The \(K\) statistic shows a considerable degree of variability but, in the absence of substantial random noise, centres slightly below \(K=1\) which suggests the statistic is behaving as expected (if not slightly conservatively) when phylogenetic signal is present. For characters whose simulated evolution is near-random, the baseline of \(K\) seems to be elevated a little by our particular reference tree, with non-phylogenetic simulated characters ranging from the expected \(K=0\) to around \(K=0.3\). This should be kept in mind when interpreting \(K\) scores across our results. As for the randomisation procedure, Figure \ref{fig:k-simulation-plots}(B) shows the percentage of simulated traits that were identified as having a significant degree of phylogenetic signal (\(p < 0.05\)) at a given level of Brownian motion mixed with random noise. Above around 65\% Brownian motion, there are no Type II errors. The ability to detect significant phylogenetic signal drops as the level of random noise increases beyond 35\%, though overall the test's sensitivity seems acceptable. At the opposite extreme, where characters are simulated completely at random (and, therefore, there is no phylogenetic signal to detect) the randomisation procedure falsely detects phylogenetic signal 5.2\% of the time, very close to the expected false discovery rate of 5\% (given the conventional threshold for statistical significance of \(\alpha = 0.05\)). On the basis of these simulations, we are satisfied that randomisation procedure is sufficiently robust, given the particular size and shape of the Pama-Nyungan reference phylogeny.

\begin{figure}

{\centering \includegraphics[width=0.66\linewidth]{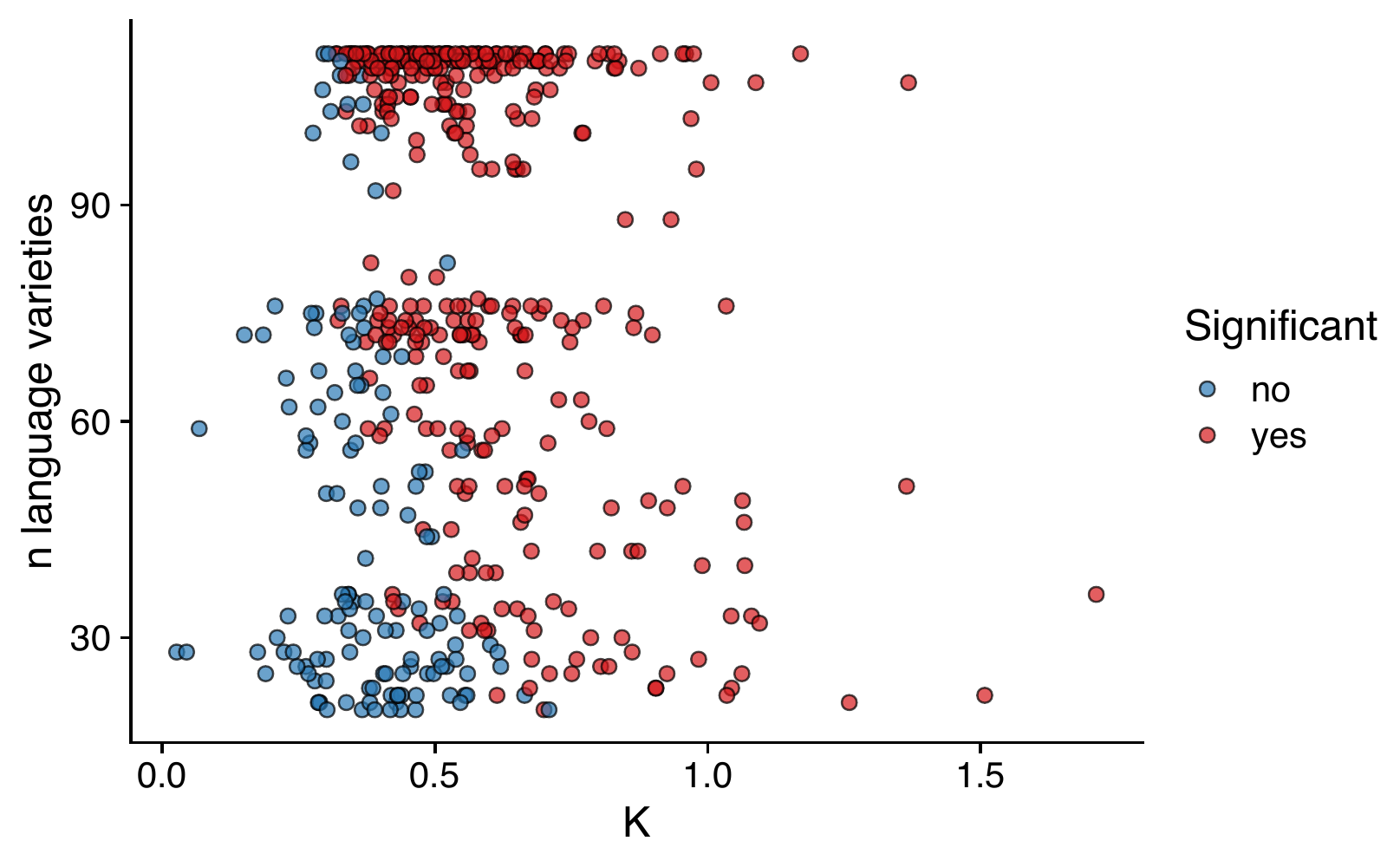} 

}

\caption{Estimates of $K$ for forward and backward transition frequency characters plotted against the number of doculects with non-missing values for each character. Although some statistical power is lost and variability increases among characters with the most missing values, $K$ scores cluster evenly around mean $K$ (0.52).}\label{fig:k-scatterplot}
\end{figure}

As a final check, we consider whether the \(K\) statistic might be affected by the quantity of missing or `not applicable' values for a given character. We inspect this visually by plotting, for all biphone characters, the relation between a biphone's \(K\) score and the number of language varieties with non-missing data points on which \(K\) was calculated (Figure \ref{fig:k-scatterplot}). When \(K\) is calculated on fewer than around 40 non-missing values, the statistic shows a wider degree of variability. In addition, phylogenetic signal is deemed statistically significant for fewer characters in this range, suggesting that the quantity of missing values is affecting the statistical power of the test. However, all \(K\) scores cluster centrally around the mean regardless of the number of languages with non-missing values, suggesting that the mean \(K\) we observe for the dataset overall is not significantly affected by missing data.

\hypertarget{fwd-vs-bkwd}{%
\subsection{Forward transitions versus backward transitions}\label{fwd-vs-bkwd}}

\begin{figure}

{\centering \includegraphics[width=0.66\linewidth]{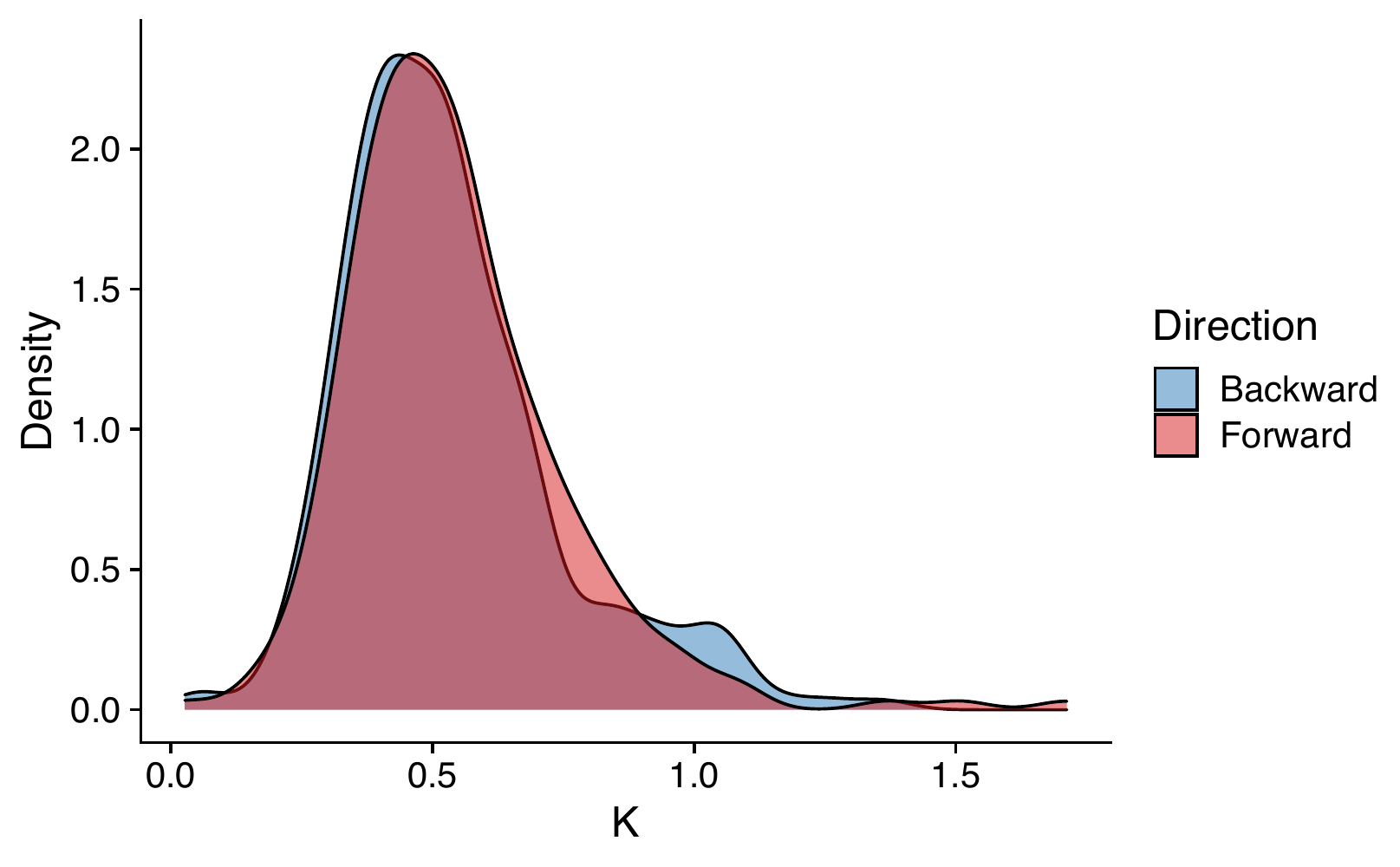} 

}

\caption{Distribution of $K$ scores for forward transition frequencies versus backward transition frequencies. We find no significant difference between these character types.}\label{fig:k-fwd-vs-bkwd}
\end{figure}

We find no significant difference in the means of \(K\) for forward transition characters (mean \(K =\) 0.54) and backward transition characters (mean \(K =\) 0.53) (\(t=\) 0.44, \(df=\) 487.98, \(p=\) 0.661, 95\% CI {[}-0.03, 0.05{]}). The distributions of \(K\) scores for forward and backward transitions are plotted in Figure \ref{fig:k-fwd-vs-bkwd}, showing a high degree of overlap between the two.

\hypertarget{norm-characters}{%
\subsection{Normalisation of character values}\label{norm-characters}}

Visual inspection of the density plots for each character shows there is a tendency for character data to be negatively skewed (the weight of the distribution is left-of-centre), although this is not universally the case. To test whether the particular, heavy-tailed nature of the data has an effect on tests for phylogenetic signal, we apply Tukey's Ladder of Powers transformation to each character in the dataset and re-run both the \(K\) test and randomization procedure. This is a power transformation, which makes the data fit a normal distribution as closely as possible. It does this by finding the power transformation value, \(\lambda\), that maximizes the \(W\) statistic of the Shapiro-Wilk test for normality for each character individually. For our purposes, this transformation is effectively a change in the evolutionary model: A Brownian motion process is still assumed---a character value may wander up or down with equal probability---but, in this model, character values shift up or down along a transformed scale.

\begin{figure}

{\centering \includegraphics[width=0.66\linewidth]{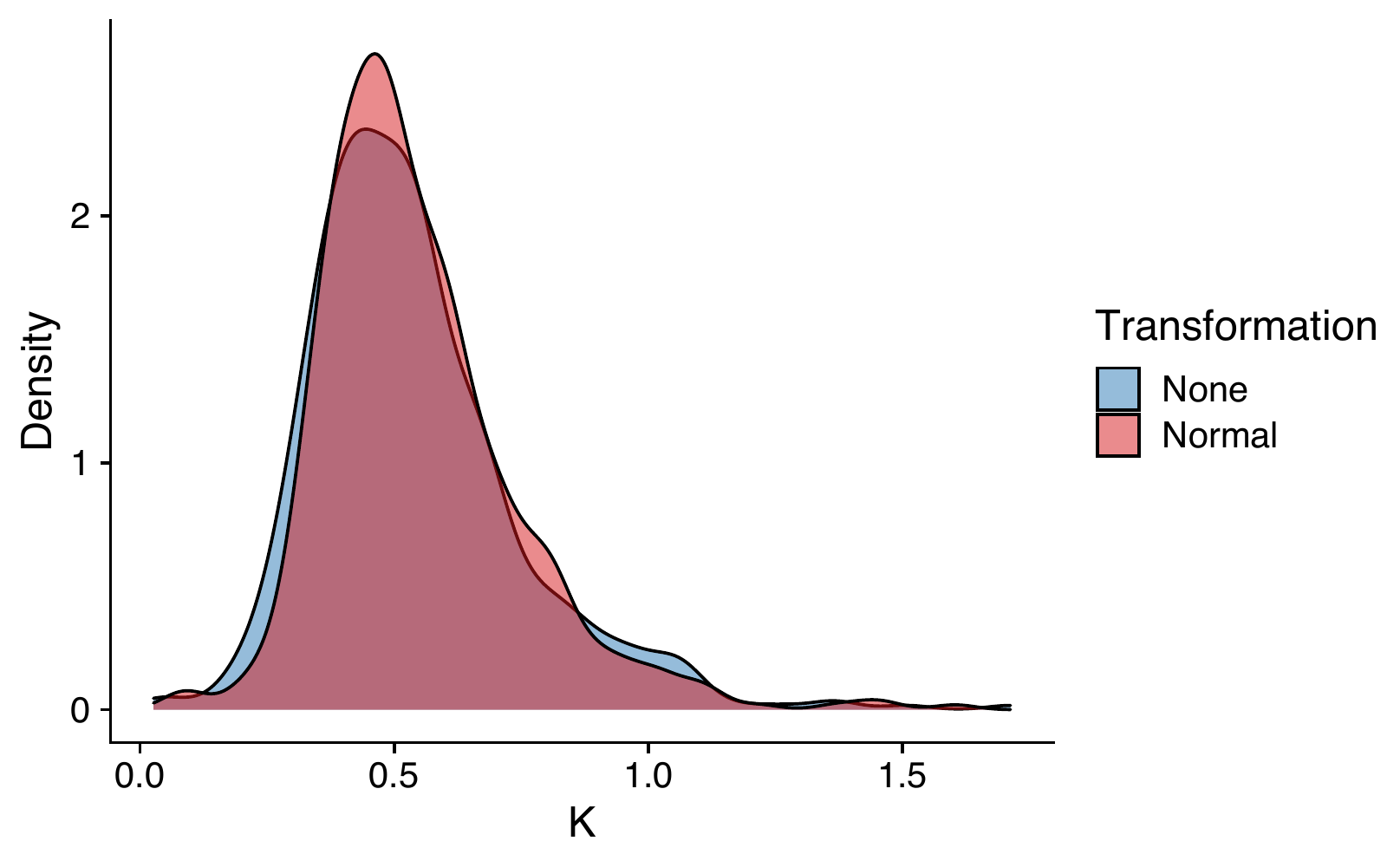} 

}

\caption{Distributions of $K$ for untransformed character values and their normalized counterparts. We find no significant difference between these distributions.}\label{fig:orig-vs-nrmlzd}
\end{figure}

Mean \(K\) for normalized character data is 0.55 (\(SD\) 0.2). Of 490 characters, 408 or 83\% (208 forward transitions, 200 backward transitions) contain phylogenetic signal significantly above the random expectation. There is no statistically significant difference between mean \(K\) for untransformed data (0.54) versus mean \(K\) for normalized data (\(t=\) 0.59, \(df=\) 973.66, \(p=\) 0.558, 95\% CI {[}-0.02, 0.03{]})---see Figure \ref{fig:orig-vs-nrmlzd}.

\hypertarget{phy-sig-classes}{%
\section{Phylogenetic signal in natural-class-based characters}\label{phy-sig-classes}}

One limitation of analysing phylogenetic signal in biphone characters is the assumption that every biphone character is a statistically independent observation. In historical linguistic processes, however, phonological segments rarely behave independently. Rather, sound changes are applied to whole sound classes, thereby affecting any one of various cross-cutting sets of phonological segments (and, therefore, biphone characters we have used in this study).

To account for this non-independence and more faithfully model what we know about how phonotactic systems operate in a language, we extract forward and backward transition probabilities for sequences of phonological features. For the purposes of this experiment, word boundaries are counted as a class and vowels are reduced to a single `vowel' class. Three sets of characters are extracted: foward and backward transition probabilities between natural classes based on place of articulation (segments belonging to the following classes: word boundary, labial, dental, alveolar, retroflex, palatal, velar, glottal, vowel); forward and backward transition probabilities between natural classes based on major places of articulation, where coronal contrasts have been collapsed (word boundary, labial, apical, laminal, velar, vowel); and natural classes based on manner of articulation (word boundary, obstruent, nasal, vibrant, lateral, glide, rhotic glide, vowel). The choice of natural classes is based on well-established principles of organisation among segments in Australian languages (Dixon \protect\hyperlink{ref-dixon_languages_1980}{1980}; Hamilton \protect\hyperlink{ref-hamilton_phonetic_1996}{1996}; Baker \protect\hyperlink{ref-baker_word_2014}{2014}; Round \protect\hyperlink{ref-round_segment_2021}{2021}\protect\hyperlink{ref-round_segment_2021}{a}; Round \protect\hyperlink{ref-round_phonotactics_2021}{2021}\protect\hyperlink{ref-round_phonotactics_2021}{b}).

\begin{table}

\caption{\label{tab:k-natural-classes-summary}Summary of $K$ analysis for forward and backward transition frequencies between different natural classes. The two rightmost columns indicate the total number of characters analysed and the percentage of those characters with a significant degree of phylogenetic signal according to the randomization procedure.}
\centering
\begin{tabular}[t]{lrrr}
\toprule
Classes & Mean K & n characters & significant (\%)\\
\midrule
Place & 0.61 & 126 & 94\\
Major place & 0.62 & 96 & 74\\
Manner & 0.59 & 88 & 66\\
\bottomrule
\end{tabular}
\end{table}

\begin{figure}

{\centering \includegraphics[width=0.66\linewidth]{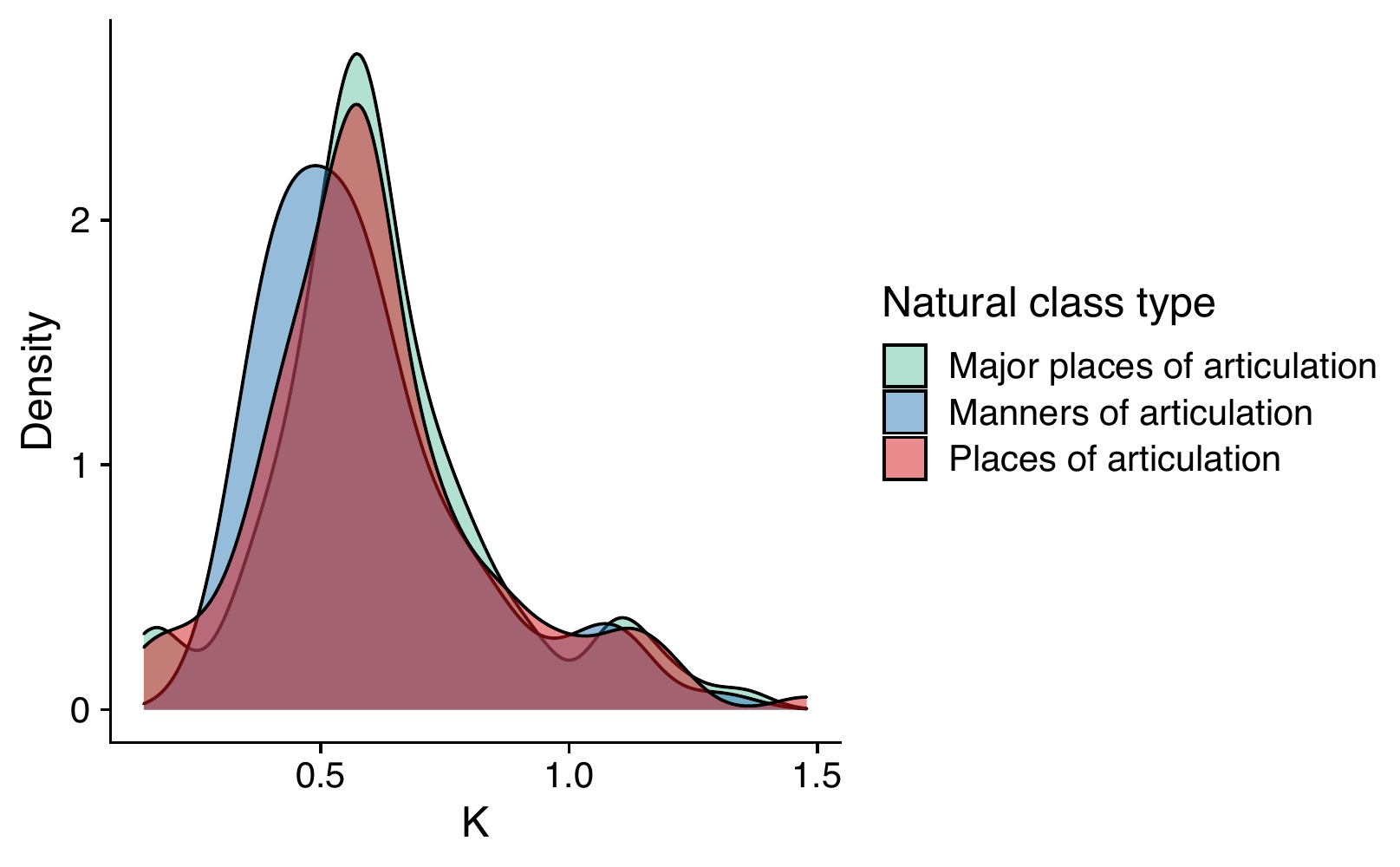} 

}

\caption{Comparison of $K$ scores for transitions between different kinds of natural classes. The differences between all three distributions are not statistically significant.}\label{fig:k-natural-classes}
\end{figure}

Table \ref{tab:k-natural-classes-summary} presents mean \(K\) and the proportion of significant characters for each of these three natural-class-based datasets. All show highly similar distributions (Figure \ref{fig:k-natural-classes}). There is no statistically significant difference in the means of \(K\) for the three feature types, according to a one-way ANOVA (\(F\)(, ) = 0.46, \(p\) = 0.6288925). An Anderson-Darling k-sample test, which tests the hypothesis that \(k\) independent samples come from a common, unspecified distribution (i.e., no prior assumption about normality) also finds no significant difference in the distributions of \(K\) scores for the three natural-class-based datasets (\(AD=\) 2.15, \(T.AD=\) 0.14, \(p=\) 0.34).

\hypertarget{classes-vs-biphones}{%
\subsection{natural-class-based characters versus biphones}\label{classes-vs-biphones}}

\begin{figure}

{\centering \includegraphics[width=0.66\linewidth]{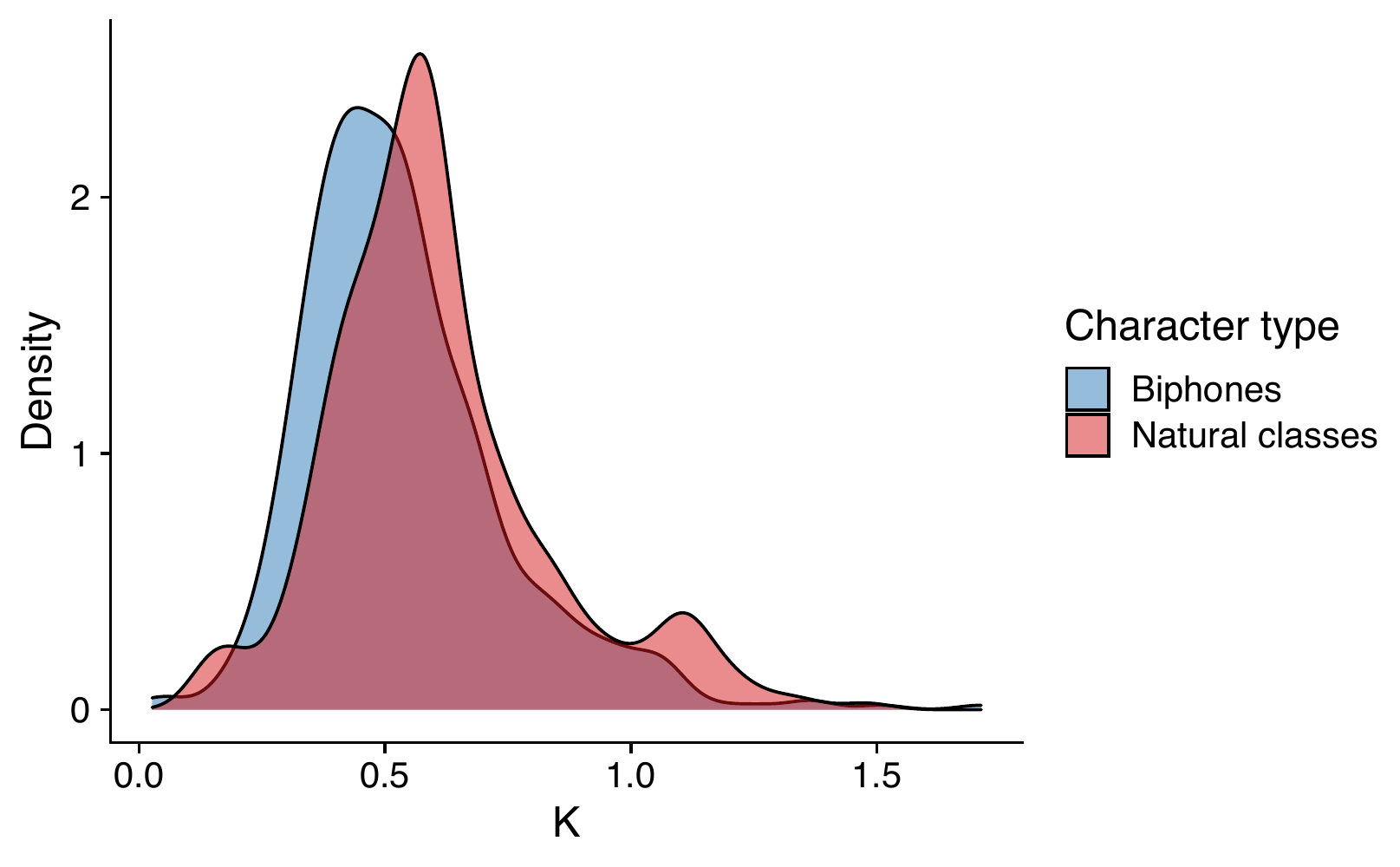} 

}

\caption{Distributions of $K$ scores for biphone characters, coding the relative frequencies of transitions between phonological segments, and natural-class-based characters, coding the relative frequencies of transitions between natural classes of segments. Phylogenetic signal is higher overall in the natural-class-based dataset than the biphone-based dataset}\label{fig:k-biphones-vs-features}
\end{figure}

We compare the degree of phylogenetic signal in the biphone data tested in Section \ref{phy-sig-cont} to the natural-class-based data tested here (see Figure \ref{fig:k-biphones-vs-features}). Mean \(K\) for all natural-class-based characters is 0.61, which is significantly higher than the mean \(K\) for biphone data, 0.54 (\(t=\) 4.37, \(df=\) 618.89, \(p=\) 0, 95\% CI {[}0.04, 0.1{]}). The Kolmogorov-Smirnov test, which is more sensitive to the overall shape of the distribution than a t-test comparison of means, also finds a significant distinction between \(K\) for biphone characters and \(K\) for natural-class-based characters (\(D=\) 0.2, \(p < 0.001\)).

\hypertarget{discussion}{%
\section{Discussion}\label{discussion}}

Computational phylogenetic methods are increasingly commonplace in historical linguistics. However, there has been relatively less consideration of the range of data types that might profitably be used with computational phylogenetic methods, beyond traditional, manually-assembled sets of lexical cognate data. In this study, we have considered the potential utility of quantitative phonotactic data for historical linguistics, for the reasons that quantitative phonotactic data is (i) readily extractible from basic wordlists, and (ii) may show certain kinds of historical conservatism, where the historical signal in more traditional lexical data would be affected by borrowing and lexical innovation.

We extracted frequencies of transitions between phonological segments in scrubbed and comparably segmented wordlists representing 111 Pama-Nyungan language varieties. As points of comparison, we extracted two additional datasets: Firstly, a binarized version of the dataset, which simply records whether or not particular two-segment sequences, a.k.a. biphones, are present in a language's wordlist. This is to emulate, in a simple sense, the kind of information which is often recorded in the phonology section of published descriptive grammars. We also extracted frequencies for transitions between natural classes of sounds. This is to account (at least, to some partial degree) for the fact that phonological segments tend not to evolve independently but pattern into natural classes, thereby limiting the independence of biphone-based variables.

To test whether historical information is preserved in our phonotactic datasets, we tested for phylogenetic signal, that is, the degree to which variance in the data reflects the evolutionary history of the 111 language varieties. We took an independent phylogenetic tree, inferred by the second author using lexical cognate data, and assumed a simple Brownian motion model of evolution. Our first key finding is that a significant degree of phylogenetic signal is detected in all three datasets---binary, segment-based and sound class-based. Finding phylogenetic signal in the binary dataset is somewhat surprising, given previous descriptions of homogeneity in the phonotactics of Australian languages. Our second key finding is that phylogenetic signal is significantly stronger in the higher-definition, frequency-based datasets than it is in binary dataset. In turn, phylogenetic signal in the sound class-based dataset is significantly stronger than the segment-based dataset. We took a closer look at certain comparisons within datasets, namely, whether there is a difference between forward and backward transitions, different kinds of sound classes, or between original (heavy-tailed) data and transformed data (to more closely fit a normal distribution). In all three cases, no significant differences were found.

\hypertarget{overall-robustness}{%
\subsection{Overall robustness}\label{overall-robustness}}

One important assumption in this study is that the tree being used as a reference phylogeny is an accurate depiction of the phylogeny for the languages in question. In the absence of time travel, it is impossible to observe directly the true Pama-Nyungan phylogeny and thus surely satisfy this assumption. Instead, we must rely on best available data and methods to infer the phylogeny. As described in Section \ref{ref-phylogeny}, the Pama-Nyungan phylogeny we use in this study was inferred by the second author using Bayesian computational phylogenetic methods, which produce a posterior sample of possible trees. The specific reference tree we use in this study is a best possible summation of this posterior sample as determined by the \emph{maximum clade credibility} method. Naturally, there is a degree of uncertainty associated with the topology of the maximum clade credibility tree. It may be the case that a better representation of the true Pama-Nyungan phylogeny exists among the many slight permutations contained within the posteior tree sample.

To evaluate the robustness of our results against phylogenetic uncertainty, we repeat the \(K\) test for phylogenetic signal on a subset of sound-class-based characters and each of a subset of 100 trees selected from the posterior sample. The subset of characters includes forward and backward transitions between place features, plus forward and backward transitions for manner features. The trees selected are the 100 best trees in the posterior sample, based on the maximum clade credibility metric. In this way, we capture a degree of uncertainty in the topology and branch lengths of the Pama-Nyungan phylogeny, while restricting our attention to a subset of the most credible alternatives.

\begin{figure}

{\centering \includegraphics[width=0.66\linewidth]{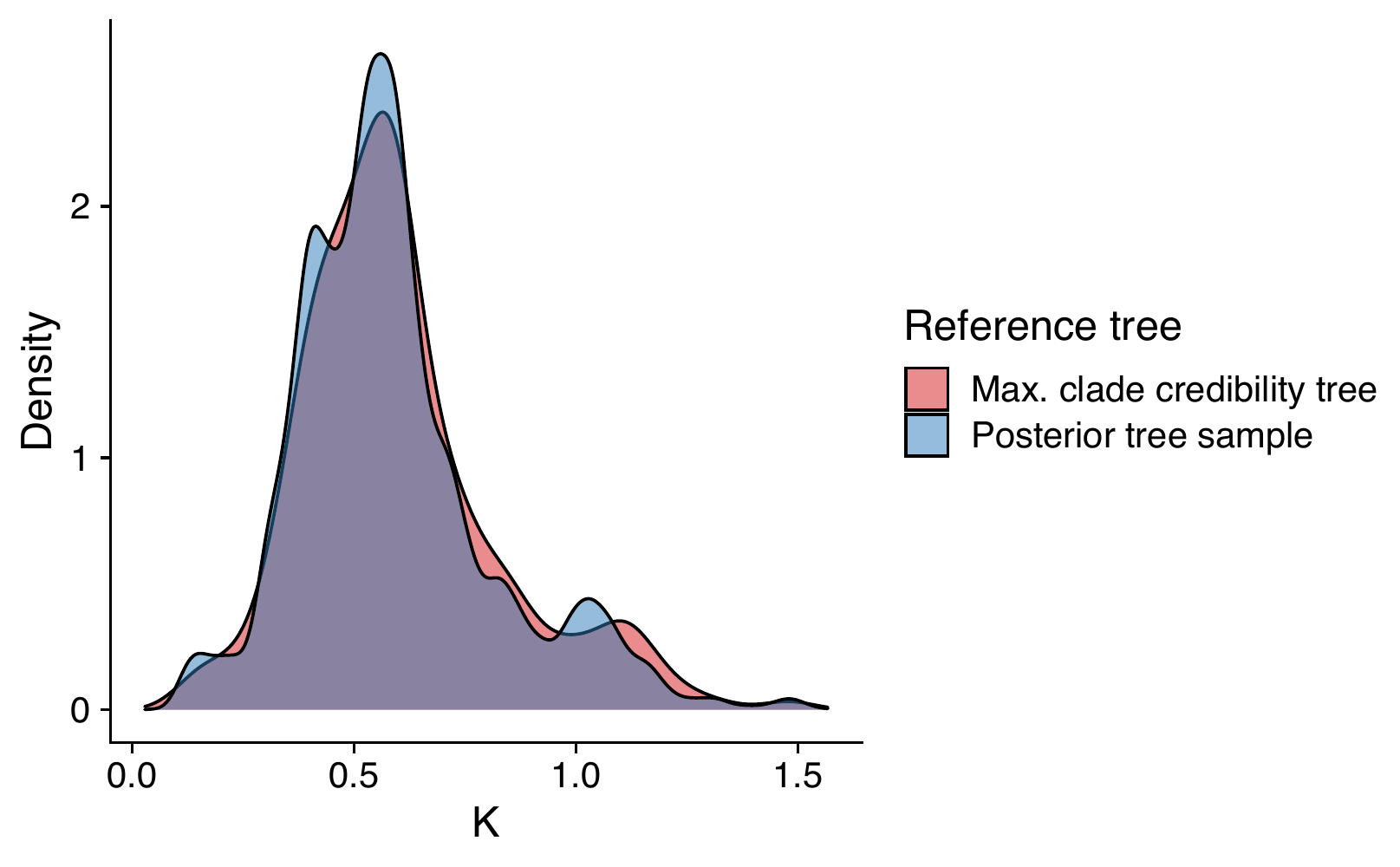} 

}

\caption{Comparison of $K$ statistics using a 100-tree posterior sample versus the maximum clade credibility tree alone.}\label{fig:tree-uncertainty}
\end{figure}

214 sound class characters in total are tested, giving 21,400 \(K\) statistics in total (each character multiplied by 100 trees). The mean of these \(K\) scores is 0.59, which compares to a mean \(K\) of 0.6 for the same characters applied only to the maximum clade credibility tree as in Section \ref{phy-sig-classes}. This difference is not statistically significant (\(t=\) -0.95, \(df=\) 217.17, \(p=\) 0.342, 95\% CI {[}-0.05, 0.02{]}). The distributions of these \(K\) scores are illustrated in Figure \ref{fig:tree-uncertainty}.

\begin{figure}

{\centering \includegraphics[width=1\linewidth]{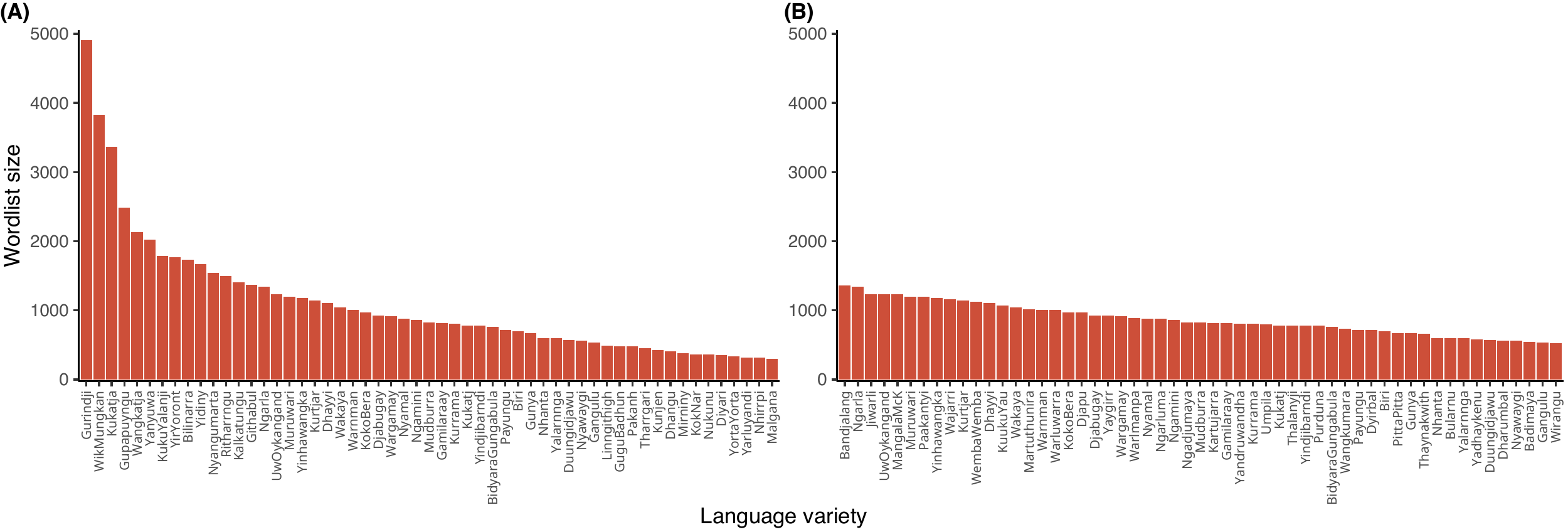} 

}

\caption{Wordlists ranked by size. (A) shows every second wordlist in the language sample. (B) shows the middle 50\% of wordlists. Each subset contains the same number of wordlists but discrepancy in their size is greatly reduce in (B).}\label{fig:wordlist-subset-sizes}
\end{figure}

\begin{figure}

{\centering \includegraphics[width=0.66\linewidth]{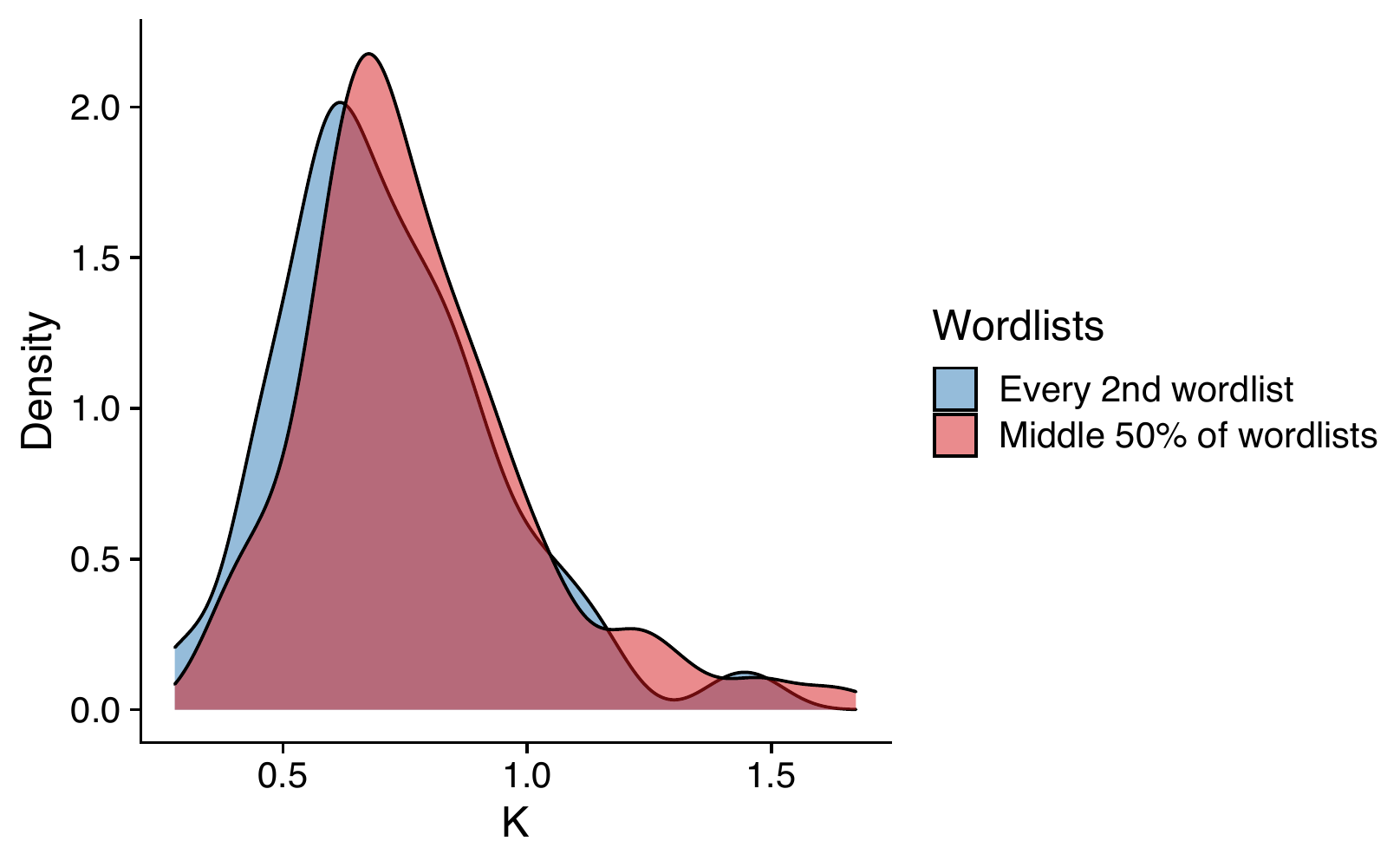} 

}

\caption{Comparison of $K$ statistics using only wordlists falling within the 25th an 75th quantiles (middle 50\%) for wordlist size, versus a sample of every 2nd wordlist (when ranked in order of size).}\label{fig:wordlist-uncertainty}
\end{figure}

In Section \ref{wordlists} we mentioned that our wordlists vary significantly is size. The length of a wordlist can correlate with many other linguistic properties that it has. For instance, in our data, the number of entries in a wordlist and the mean phonemic length of those entries have a Pearson's correlation coefficient of \(r = 0.71\) (\(p < 0.001\)). That is, longer lists tend to contain longer words (presumably since shorter lists are weighted towards more basic, shorter vocabulary items). The existence of correlations like this means that it is not possible simply to `counterbalance' the length of wordlists by sub-sampling or resampling their items so that the resampled lists all have the same length. For example, the resampled lists that derive from longer underlying lists would still contain longer words. Nevertheless, it would still be desirable to know whether our results in Sections \ref{phy-sig-bin}--\ref{phy-sig-classes} are unduly influenced by the disparity in our wordlist lengths, by manipulating it in a controlled fashion. To do this, we extracted two different language sub-samples from our dataset. For the first, we ranked all wordlists by size and selected every second language, producing a sample half the original size but with the same disparity in lengths. For the second, we ranked all wordlists by size and selected the middle 50\% of the ranking, again producing a sample of half the size but this time with heavily reduced disparity, as shown in Figure \ref{fig:wordlist-subset-sizes}. Wordlists in the middle 50\% range in length from 528 to 1361 (mean 872).

For these samples, we ran \(K\) tests on place and manner natural class characters. Our reasoning is that if wordlist disparity strongly affects the estimation of phylogenetic signal, then we should see a clear difference in the results. Mean \(K\) for the middle 50\% of wordlists is 0.77, which is somewhat greater than the mean for every second wordlist 0.72. This difference is small but statistically significant (\(t=\) 2.44, \(df=\) 423.93, \(p=\) 0.015, 95\% CI {[}0.01, 0.1{]}). This suggests that disparities in wordlist length are somewhat degrading the phylogenetic signal in our data, although there remains broad similarity between them, as pictured in Figure \ref{fig:wordlist-uncertainty}.

One attributing factor for this small degradation in phylogenetic signal may be measurement error among the bottom quartile of small wordlists. Throughout this study, we have assumed that all character values are accurate and do not account for measurement error. Accounting for measurement error when testing for phylogenetic signal is an area of active development in comparative biology (Zheng et al. \protect\hyperlink{ref-zheng_new_2009}{2009}). In future studies, this degradation in phylogenetic signal could be investigated by relating variation in \(K\) statistics carefully to various linguistic properties that correlate with the length of wordlists.

\hypertarget{limitations}{%
\subsection{Limitations}\label{limitations}}

Any investigation of phylogenetic signal in essence is an investigation of cross-linguistic (dis)similarity. Accordingly, whenever our representations of linguistic facts are altered, then the phylogenetic signal detectable in them will almost certainly change to some degree. In this paper our focus has been trained on the initial, fundamental question of whether phylogenetic signal is detectable in maximally simple phonotactic characters. However, as work like ours increases, one priority will be to investigate how investigators' choices about how data is represented affects results.

Relevant for the current study, in Section \ref{wordlists} we described a process of normalisation. The motivation for this was to attempt to minimize certain aspects of variation in phonological representation that can arise from variation in how different linguists analyses the same essential facts (Chao \protect\hyperlink{ref-chao_non-uniqueness_1934}{1934}; Hockett \protect\hyperlink{ref-hockett_problem_1963}{1963}; Hyman \protect\hyperlink{ref-hyman_universals_2008}{2008}; Dresher \protect\hyperlink{ref-dresher_contrastive_2009}{2009}). While normalisation per se ought to improve the quality of cross-linguistic comparisons that the data enables, there is still the question of which targets one ought to normalize the data towards, on what effect that choice can have. For example, a reviewer asks whether our choice to split up complex segments might amplify phylogenetic signal if it leads to certain phylogenetically distributed complex segments counting instead as biphones. This can be answered in three ways. First, in the general case, since splitting segments changes representations, it will alter aspects of (dis)similarity in the data, and so is very likely to affect phylogenetic signal in some manner. Second, in this particular case, the reviewer is likely to be correct, due to details of our method. Any cross-linguistically rare, complex segment would likely get excluded from our dataset. This is because, although it would figure in certain biphones, we have made use only of biphones that reach a minimal level of recurrence across our language sample, and thus the biphones containing the rare segment quite likely would not qualify. However, if we split this complex segment into two segments, thus into a biphone, the resulting biphone may well have sufficient cross-linguistic recurrence to qualify for inclusion in our dataset, and subsequently may contribute to raising phylogenetic signal. A final observation on this point is that such questions, about how choices in data representation interact with the results from corresponding quantitative analysis, are made tractable by our method of data preparation. Unlike many state-of-the-art cross-linguistic datasets, in which values for each language are hand-coded and thus incapable of being `recalculated' under altered assumptions, our phonotactic characters are generated algorithmically from an underlying, very rich dataset. With a change to algorithmic parameters, we can systematically split segments or glue them together, neutralize them or keep them distinct, and document what we have done and how. As mentioned, in this paper our focus is on the simple existence of phylogenetic signal. Our methods, though, naturally extend to enable comprehensive checking of such interactions between data choices and results. Ultimately, as a discipline, we would like this to be true for all typological research, not just phylogenetics (Round \protect\hyperlink{ref-round_matthew_2017}{2017}\protect\hyperlink{ref-round_matthew_2017}{a}). An advantage of our general approach, is that it open the doors to this rigorous mode of inquiry.

A further limitation of this study relates to the assumption that the data being tested for phylogenetic signal are independent of the data that was used to infer the reference phylogeny. In this study, the wordlists from which we extracted phonotactic characters contain, as a small subset, the basic vocabulary items from which lexical cognate characters were inferred and subsequently used to build the reference phylogeny. It is unclear exactly to what degree this inclusion of basic vocabulary compromizes the independence of our reference tree and phonotactic data. A reviewer points out that cognate data and phonotactic data are still somewhat independent, even when extracted from identical wordlists, since phonotactic attributes are not directly called on to make cognacy judgements. Nevertheless, sound change affects both phonotactics and cognate identification, so some degree of non-indepenence is to be expected. To ascertain whether this effect is significant, future studies could parameterize the inclusion/exclusion of basic vocabulary from the phonotactic data.

One reviewer raises the correlation between phylogeny and geography. A noted limitation of phylogenetic comparative methods is the inability to account for geography as a possible confound (Sookias, Passmore \& Atkinson \protect\hyperlink{ref-sookias_deep_2018}{2018}) and this limitation applies to this study. Although we leave it as a priority for future work, the task of disentangling phylogeny and geography is not intractable. For example, Freckleton \& Jetz (\protect\hyperlink{ref-freckleton_space_2009}{2009}) present a method for quantifying the relative degree of spatial versus phylogenetic effects in comparative characters.

One final point to note is that recent research suggests that phylogenetic signal can be inflated when character values evolve according to a Lévy process, where a character value can wander as per a Brownian motion process, but with the addition of discontinuous paths (i.e., sudden jumps in the character's value) (Uyeda et al. \protect\hyperlink{ref-uyeda_rethinking_2018}{2018}). This is a realistic concern in the linguistic context, where segment frequencies are subject to sudden shifts caused by phonological mergers and splits. The possibility of Lévy-like evolutionary processes is a matter of concern also in comparative biology and methods to investigate it are subject to active development in that field (Uyeda et al. \protect\hyperlink{ref-uyeda_rethinking_2018}{2018}).

\hypertarget{conclusion}{%
\section{Conclusion}\label{conclusion}}

Historical and synchronic comparative linguistics are increasingly making use of phylogenetic methods for the same reasons that led biologist to switch to them several decades ago. Our central contention has been that phylogenetic methods not only give us new ways of studying existing comparative data sets, but open up the possibility to derive insights from new kinds of data. Here we demonstrate the potential for phylogenetically investigating phonotactic data, by showing that it indeed contains the kind of phylogenetic signal which is the prerequisite for a whole spectrum of phylogenetic analyses.

We find significant phylogenetic signal for several hundred phonotactic characters extracted semi-automatically from 111 Pama-Nyungan wordlists, demonstrating that historical information is detectable in phonotactic data, even at the relatively simple level of biphones and despite ostensibly high phonological uniformity. Contrary to the prevailing view in literature on Australian languages, and contrary to the findings of an earlier pilot study on a much smaller language sample, we find that binary characters marking the presence or absence of biphones in a doculect contain enough phylogenetically-patterned variation to detect phylogenetic signal. However, we find that statistical power is relatively low when operating with coarse-grained binary data and quantification of the degree of phylogenetic signal is affected by a large number of low-variation characters, where all but one or a few doculects share the same value. We find stronger phylogenetic signal in biphone characters of forward and backward transition frequencies. This reaffirms the results of earlier work, for the first time on a sample of languages spanning an entire large family and the vast majority of a continent. It also reaffirms earlier findings that Australian phonologies show a greater level of variation than traditionally has been appreciated, once matters of frequency are taken into account. We find a significantly greater level of phylogenetic signal again in characters based on the frequencies of forward and backward transitions between natural sound classes. The sound-class-based approach reduces the quantity of characters available to test, but limits sparsity in the dataset and accounts somewhat better for the role of sound-classes in the evolutionary processes that affect phonotactic patterns in human language. Interestingly, although there exists considerable variation in the level of phylogenetic signal found in individual characters, we find no observable pattern to this variation in segment-based biphone characters nor between the mean levels of phylogenetic signal observed for different kinds of sound classes (e.g., characters concerning place versus manner of articulation).

This work has implications for comparative linguistics, both typological and historical. Firstly, we recommend the use of phylogenetic comparative methods in typological work where the phylogenetic independence of a language sample (or lack thereof) is paramount. In the immediate term, this should be the case for any typological work concerning phonotactics, even in parts of the world such as Australia where phonotactics traditionally have been assumed to be relatively independent of phylogeny. Beyond phonotactics, however, explicit measurements of phylogenetic signal can be made for any set of cross-linguistic data and this can be built into statistical analysis, even in the presence of gaps and uncertainty in phylogenetic knowledge. In two decades of quantitative development in historical linguistics, there has still been relatively limited consideration of the kinds of characters used for inferring linguistic histories. The phonotactic characters presented here can be extracted relatively simply and in large quantities from wordlists, even where a full descriptive grammar is not available. Here, we test only the degree to which patterns of variation in our data match our independent, pre-existing knowledge of the phylogenetic history of the Pama-Nyungan family, however, the results suggest that phonotactic data of this kind could be used where the phylogeny is less certain, either by incorporating phonotactic data into phylogenetic inference directly or by constraining parts of the tree where lexical data on its own returns some doubt.

\hypertarget{acknowledgements}{%
\section*{Acknowledgements}\label{acknowledgements}}
\addcontentsline{toc}{section}{Acknowledgements}

JM-C is supported by an Australian Government Research Training Program Scholarship. ER is supported by Australian Research Council (ARC) Discovery Grant DE150101024 and the grant \emph{Auto-harvested insights from wordlists} from the ARC Centre of Excellence for the Dynamics of Language. CB is supported by National Science Foundation Grant NSF1423711. We gratefully acknowledge this support.

\hypertarget{references}{%
\section*{References}\label{references}}
\addcontentsline{toc}{section}{References}

\hypertarget{refs}{}
\leavevmode\hypertarget{ref-albright_rules_2003}{}%
Albright, Adam \& Bruce Hayes. 2003. Rules vs. Analogy in English past tenses: A computational/experimental study. \emph{Cognition} 90(2). 119--161. doi:\href{https://doi.org/10.1016/S0010-0277(03)00146-X}{10.1016/S0010-0277(03)00146-X}.

\leavevmode\hypertarget{ref-alpher_pama-nyungan:_2004}{}%
Alpher, Barry Jacob. 2004. Pama-Nyungan: Phonological reconstruction and status as a phylogenetic group. In Claire Bowern \& Harold Koch (eds.), \emph{Australian languages: Classification and the comparative method} (Current Issues in Linguistic Theory 249), 93--126. Amsterdam: John Benjamins Publishing Company.

\leavevmode\hypertarget{ref-austin_proto-kanyara_1981}{}%
Austin, Peter. 1981. Proto-Kanyara and Proto-Mantharta historical phonology. \emph{Lingua} 54(4). 295--333. doi:\href{https://doi.org/10.1016/0024-3841(81)90009-7}{10.1016/0024-3841(81)90009-7}.

\leavevmode\hypertarget{ref-baker_word_2014}{}%
Baker, Brett. 2014. Word structure in Australian languages. In, \emph{The languages and linguistics of Australia: A comprehensive guide}, 139--214. Berlin: De Gruyter Mouton. doi:\href{https://doi.org/10.1515/9783110279771.139}{10.1515/9783110279771.139}.

\leavevmode\hypertarget{ref-balisi_dietary_2018}{}%
Balisi, Mairin, Corinna Casey \& Blaire Van Valkenburgh. 2018. Dietary specialization is linked to reduced species durations in North American fossil canids. \emph{Royal Society Open Science} 5(4). 171861. doi:\href{https://doi.org/10.1098/rsos.171861}{10.1098/rsos.171861}.

\leavevmode\hypertarget{ref-birchall_comparison_2015}{}%
Birchall, Joshua. 2015. A comparison of verbal person marking across Tupian languages. \emph{Boletim do Museu Paraense Emílio Goeldi. Ciências Humanas} 10(2). 325--345. doi:\href{https://doi.org/10.1590/1981-81222015000200007.}{10.1590/1981-81222015000200007.}

\leavevmode\hypertarget{ref-blasi_human_2019}{}%
Blasi, Damián E., Steven Moran, Scott R. Moisik, Paul Widmer, Dan Dediu \& Balthasar Bickel. 2019. Human sound systems are shaped by post-Neolithic changes in bite configuration. \emph{Science} 363(6432). doi:\href{https://doi.org/10.1126/science.aav3218}{10.1126/science.aav3218}.

\leavevmode\hypertarget{ref-blomberg_tempo_2002}{}%
Blomberg, Simon P. \& Theodore Garland Jr. 2002. Tempo and mode in evolution: Phylogenetic inertia, adaptation and comparative methods. \emph{Journal of Evolutionary Biology} 15(6). 899--910. doi:\href{https://doi.org/10.1046/j.1420-9101.2002.00472.x}{10.1046/j.1420-9101.2002.00472.x}.

\leavevmode\hypertarget{ref-blomberg_testing_2003}{}%
Blomberg, Simon P., Theodore Garland Jr. \& Anthony R. Ives. 2003. Testing for phylogenetic signal in comparative data: Behavioral traits are more labile. \emph{Evolution} 57(4). 717--745.

\leavevmode\hypertarget{ref-bouckaert_origin_2018}{}%
Bouckaert, Remco R., Claire Bowern \& Quentin D. Atkinson. 2018. The origin and expansion of Pama-Nyungan languages across Australia. \emph{Nature Ecology \& Evolution} 2(4). 741--749. doi:\href{https://doi.org/10.1038/s41559-018-0489-3}{10.1038/s41559-018-0489-3}.

\leavevmode\hypertarget{ref-bowern_pama-nyungan_2015}{}%
Bowern, Claire. 2015. Pama-Nyungan phylogenetics and beyond {[}plenary address{]}. In, \emph{Lorentz center workshop on phylogenetic methods in linguistics}. Leiden University, Leiden, Netherlands. doi:\href{https://doi.org/10.5281/zenodo.3032846}{10.5281/zenodo.3032846}.

\leavevmode\hypertarget{ref-bowern_chirila:_2016}{}%
Bowern, Claire. 2016. Chirila: Contemporary and historical resources for the indigenous languages of Australia. \emph{Language Documentation and Conservation} 10. \url{http://hdl.handle.net/10125/24685}.

\leavevmode\hypertarget{ref-bowern_standard_2017}{}%
Bowern, Claire. 2017. Standard Average Australian? In, \emph{12th conference of the association for linguistic typology (ALT 2017)}. Australian National University, Canberra, Australia. doi:\href{https://doi.org/10.5281/zenodo.1104222}{10.5281/zenodo.1104222}.

\leavevmode\hypertarget{ref-bowern_computational_2018}{}%
Bowern, Claire. 2018a. Computational phylogenetics. \emph{Annual Review of Linguistics} 4(1). 281--296. doi:\href{https://doi.org/10.1146/annurev-linguistics-011516-034142}{10.1146/annurev-linguistics-011516-034142}.

\leavevmode\hypertarget{ref-bowern_pama-nyungan_2018}{}%
Bowern, Claire. 2018b. Pama-Nyungan cognate judgements: 285 languages. \url{10.5281/zenodo.1318310}.

\leavevmode\hypertarget{ref-bowern_computational_2012}{}%
Bowern, Claire \& Quentin D. Atkinson. 2012. Computational phylogenetics and the internal structure of Pama-Nyungan. \emph{Language} 88(4). 817--845. doi:\href{https://doi.org/10.1353/lan.2012.0081}{10.1353/lan.2012.0081}.

\leavevmode\hypertarget{ref-bowern_does_2011}{}%
Bowern, Claire, Patience Epps, Russell Gray, Jane Hill, Keith Hunley, Patrick McConvell \& Jason Zentz. 2011. Does lateral transmission obscure inheritance in hunter-gatherer languages? \emph{PLoS ONE} 6(9). e25195. doi:\href{https://doi.org/10.1371/journal.pone.0025195}{10.1371/journal.pone.0025195}.

\leavevmode\hypertarget{ref-bowern_australian_2004}{}%
Bowern, Claire \& Harold Koch (eds.). 2004. \emph{Australian languages: Classification and the comparative method} (Current Issues in Linguistic Theory 249). Amsterdam: John Benjamins.

\leavevmode\hypertarget{ref-busby_distribution_1982}{}%
Busby, Peter A. 1982. The distribution of phonemes in Australian Aboriginal languages. \emph{Pacific Linguistics. Series A.} (60). 73--139.

\leavevmode\hypertarget{ref-calude_typology_2016}{}%
Calude, Andreea S. \& Annemarie Verkerk. 2016. The typology and diachrony of higher numerals in Indo-European: A phylogenetic comparative study. \emph{Journal of Language Evolution} 1(2). 91--108. doi:\href{https://doi.org/10.1093/jole/lzw003}{10.1093/jole/lzw003}.

\leavevmode\hypertarget{ref-campbell_historical_2004}{}%
Campbell, Lyle. 2004. \emph{Historical linguistics: An introduction}. Cambridge, MA: MIT Press.

\leavevmode\hypertarget{ref-capell_new_1956}{}%
Capell, Arthur. 1956. \emph{A new approach to Australian linguistics}. Sydney: University of Sydney.

\leavevmode\hypertarget{ref-chang_ancestry-constrained_2015}{}%
Chang, Will, Chundra Cathcart, David Hall \& Andrew Garrett. 2015. Ancestry-constrained phylogenetic analysis supports the Indo-European steppe hypothesis. \emph{Language} 91(1). 194--244. doi:\href{https://doi.org/10.1353/lan.2015.0005}{10.1353/lan.2015.0005}.

\leavevmode\hypertarget{ref-chao_non-uniqueness_1934}{}%
Chao, Yuen-Ren. 1934. The non-uniqueness of phonemic solutions of phonetic systems. \emph{Bulletin of the Institute of History and Philology, Academia Sinica} 4(4). 363--397.

\leavevmode\hypertarget{ref-coleman_stochastic_1997}{}%
Coleman, John \& Janet Pierrehumbert. 1997. Stochastic phonological grammars and acceptability. In, \emph{Computational phonology: The third meeting of the ACL special interest group in computational phonology}, 49--56. Somerset, NJ: Association for Computational Linguistics. \url{http://arxiv.org/abs/cmp-lg/9707017} (8 March, 2018).

\leavevmode\hypertarget{ref-crawford_adaptation_2009}{}%
Crawford, Clifford James. 2009. \emph{Adaptation and transmission in Japanese loanword phonology}. Cornell University thesis. \url{http://core.ac.uk/download/pdf/4912071.pdf}.

\leavevmode\hypertarget{ref-cysouw_towards_2007}{}%
Cysouw, Michael \& Jeff Good. 2007. Towards a comprehensive languoid catalog. In, \emph{Language catalogue meeting}. Leipsig, Germany. \url{http://cysouw.de/home/presentations_files/cysouwCATALOGUE_slides.pdf} (11 December, 2019).

\leavevmode\hypertarget{ref-delsuc_phylogenomics_2005}{}%
Delsuc, Frédéric, Henner Brinkmann \& Hervé Philippe. 2005. Phylogenomics and the reconstruction of the tree of life. \emph{Nature Reviews Genetics} 6(5). 361--375. doi:\href{https://doi.org/10.1038/nrg1603}{10.1038/nrg1603}.

\leavevmode\hypertarget{ref-dixon_languages_1980}{}%
Dixon, R. M. W. 1980. \emph{The languages of Australia} (Cambridge Language Surveys). Cambridge: Cambridge University Press.

\leavevmode\hypertarget{ref-dockum_phylogeny_2018}{}%
Dockum, Rikker. 2018. Phylogeny in phonology: How tai sound systems encode their past. In, \emph{Supplemental proceedings of the 2017 annual meeting on phonology}. New York: Linguistic Society of America. doi:\href{https://doi.org/10.3765/amp.v5i0.4238}{10.3765/amp.v5i0.4238}.

\leavevmode\hypertarget{ref-dockum_swadesh_2019}{}%
Dockum, Rikker \& Claire Bowern. 2019. Swadesh lists are not long enough: Drawing phonological generalizations from limited data. \emph{Language Documentation and Description} 16. 35--54. \url{http://www.elpublishing.org/PID/168}.

\leavevmode\hypertarget{ref-dresher_contrastive_2009}{}%
Dresher, B. Elan. 2009. \emph{The contrastive hierarchy in phonology}. Cambridge: Cambridge University Press.

\leavevmode\hypertarget{ref-dresher_main_2005}{}%
Dresher, B. Elan \& Aditi Lahiri. 2005. Main stress left in early Middle English. In Michael Fortescue, Jens Erik Mogensen \& Lene Schøsler (eds.), \emph{Historical linguistics 2003: Selected papers from the 16th international conference on historical linguistics}, 76--85. Amsterdam: John Benjamins. doi:\href{https://doi.org/10.1075/cilt.257}{10.1075/cilt.257}.

\leavevmode\hypertarget{ref-dunn_dative_2017}{}%
Dunn, Michael, Tonya Kim Dewey, Carlee Arnett, Thórhallur Eythórsson \& Jóhanna Barðdal. 2017. Dative sickness: A phylogenetic analysis of argument structure evolution in Germanic. \emph{Language} 93(1). e1--e22. doi:\href{https://doi.org/10.1353/lan.2017.0012}{10.1353/lan.2017.0012}.

\leavevmode\hypertarget{ref-dunn_evolved_2011}{}%
Dunn, Michael, Simon J. Greenhill, Stephen C. Levinson \& Russell D. Gray. 2011. Evolved structure of language shows lineage-specific trends in word-order universals. \emph{Nature} 473(7345). 79--82. doi:\href{https://doi.org/10.1038/nature09923}{10.1038/nature09923}.

\leavevmode\hypertarget{ref-dunn_structural_2005}{}%
Dunn, Michael, Angela Terrill, Ger Reesink, Robert A. Foley \& Stephen C. Levinson. 2005. Structural phylogenetics and the reconstruction of ancient language history. \emph{Science} 309(5743). 2072--2075. doi:\href{https://doi.org/10.1126/science.1114615}{10.1126/science.1114615}.

\leavevmode\hypertarget{ref-eddington_spanish_2004}{}%
Eddington, David. 2004. \emph{Spanish phonology and morphology: Experimental and quantitative perspectives}. Amsterdam: John Benjamins.

\leavevmode\hypertarget{ref-ernestus_predicting_2003}{}%
Ernestus, Mirjam Theresia Constantia \& R. Harald Baayen. 2003. Predicting the unpredictable: Interpreting neutralized segments in dutch. \emph{Language} 79(1). 5--38. doi:\href{https://doi.org/10.1353/lan.2003.0076}{10.1353/lan.2003.0076}.

\leavevmode\hypertarget{ref-felsenstein_phylogenies_1985}{}%
Felsenstein, Joseph. 1985. Phylogenies and the comparative method. \emph{The American Naturalist} 125(1). 1--15. doi:\href{https://doi.org/10.1086/284325}{10.1086/284325}.

\leavevmode\hypertarget{ref-freckleton_phylogenetic_2002}{}%
Freckleton, Robert P., Paul H. Harvey \& Mark Pagel. 2002. Phylogenetic analysis and comparative data: A test and review of evidence. \emph{The American Naturalist} 160(6). 712--726. doi:\href{https://doi.org/10.1086/343873}{10.1086/343873}.

\leavevmode\hypertarget{ref-freckleton_space_2009}{}%
Freckleton, Robert P. \& Walter Jetz. 2009. Space versus phylogeny: Disentangling phylogenetic and spatial signals in comparative data. \emph{Proceedings of the Royal Society B: Biological Sciences} 276(1654). 21--30. doi:\href{https://doi.org/10.1098/rspb.2008.0905}{10.1098/rspb.2008.0905}. \url{https://royalsocietypublishing.org/doi/10.1098/rspb.2008.0905}.

\leavevmode\hypertarget{ref-fritz_selectivity_2010}{}%
Fritz, Susanne A. \& Andy Purvis. 2010. Selectivity in mammalian extinction risk and threat types: A new measure of phylogenetic signal strength in binary traits. \emph{Conservation Biology} 24(4). 1042--1051. doi:\href{https://doi.org/10.1111/j.1523-1739.2010.01455.x}{10.1111/j.1523-1739.2010.01455.x}.

\leavevmode\hypertarget{ref-garland_jr._polytomies_1999}{}%
Garland, Theodore, Jr. \& Ramón Díaz-Uriarte. 1999. Polytomies and phylogenetically independent contrasts: Examination of the bounded degrees of freedom approach. \emph{Systematic Biology} 48(3). 547--558. doi:\href{https://doi.org/10.1080/106351599260139}{10.1080/106351599260139}.

\leavevmode\hypertarget{ref-gasser_revisiting_2014}{}%
Gasser, Emily \& Claire Bowern. 2014. Revisiting phonotactic generalizations in Australian languages. In, \emph{Supplemental proceedings of the 2013 annual meeting on phonology}. University of Massachusetts, Amherst: Linguistic Society of America. doi:\href{https://doi.org/10.3765/amp.v1i1.17}{10.3765/amp.v1i1.17}.

\leavevmode\hypertarget{ref-good_languoid_2013}{}%
Good, Jeff \& Michael Cysouw. 2013. Languoid, doculect, and glossonym: Formalizing the notion 'language'. \emph{Language Documentation and Conservation} 7. 331--359. \url{http://hdl.handle.net/10125/4606}.

\leavevmode\hypertarget{ref-gordon_phonological_2016}{}%
Gordon, Matthew K. 2016. \emph{Phonological typology} (Oxford Surveys in Phonology and Phonetics 1). Oxford: Oxford University Press.

\leavevmode\hypertarget{ref-grafen_phylogenetic_1989}{}%
Grafen, Alan. 1989. The phylogenetic regression. \emph{Philosophical Transactions of the Royal Society of London. Series B, Biological Sciences} 326(1233). 119--157. doi:\href{https://doi.org/10.1098/rstb.1989.0106}{10.1098/rstb.1989.0106}.

\leavevmode\hypertarget{ref-greenhill_does_2009}{}%
Greenhill, Simon J., Thomas E. Currie \& Russell D. Gray. 2009. Does horizontal transmission invalidate cultural phylogenies? \emph{Proceedings of the Royal Society of London B: Biological Sciences} 276(1665). 2299--2306. doi:\href{https://doi.org/10.1098/rspb.2008.1944}{10.1098/rspb.2008.1944}.

\leavevmode\hypertarget{ref-greenhill_evolutionary_2017}{}%
Greenhill, Simon J., Chieh-Hsi Wu, Xia Hua, Michael Dunn, Stephen C. Levinson \& Russell D. Gray. 2017. Evolutionary dynamics of language systems. \emph{Proceedings of the National Academy of Sciences} 114(42). E8822--E8829. doi:\href{https://doi.org/10.1073/pnas.1700388114}{10.1073/pnas.1700388114}.

\leavevmode\hypertarget{ref-hamilton_phonetic_1996}{}%
Hamilton, Philip James. 1996. \emph{Phonetic constraints and markedness in the phonotactics of Australian Aboriginal languages}. Toronto: University of Toronto thesis.

\leavevmode\hypertarget{ref-hayes_stochastic_2006}{}%
Hayes, Bruce \& Zsuzsa Cziráky Londe. 2006. Stochastic phonological knowledge: The case of Hungarian vowel harmony. \emph{Phonology} 23(1). 59--104. doi:\href{https://doi.org/10.1017/S0952675706000765}{10.1017/S0952675706000765}.

\leavevmode\hypertarget{ref-hockett_problem_1963}{}%
Hockett, Charles F. 1963. The problem of universals in language. In Joseph Greenberg (ed.), \emph{Universals of language}, 1--29. Cambridge, MA: MIT Press.

\leavevmode\hypertarget{ref-hutchinson_contemporary_2018}{}%
Hutchinson, Matthew C., Marília P. Gaiarsa \& Daniel B. Stouffer. 2018. Contemporary ecological interactions improve models of past trait evolution. \emph{Systematic Biology} 67(5). 861--872. doi:\href{https://doi.org/10.1093/sysbio/syy012}{10.1093/sysbio/syy012}.

\leavevmode\hypertarget{ref-hyman_role_1970}{}%
Hyman, Larry M. 1970. The role of borrowing in the justification of phonological grammars. \emph{Studies in African Linguistics} 1(1). 1--48. \url{https://journals.linguisticsociety.org/elanguage/sal/article/view/927.html}.

\leavevmode\hypertarget{ref-hyman_universals_2008}{}%
Hyman, Larry M. 2008. Universals in phonology. \emph{The Linguistic Review} 25(1). 83--137. doi:\href{https://doi.org/10.1515/TLIR.2008.003}{10.1515/TLIR.2008.003}.

\leavevmode\hypertarget{ref-kang_loanword_2011}{}%
Kang, Yoonjung. 2011. Loanword phonology. In Marc Van Oostendorp, Colin J. Ewen, Elizabeth Hume \& Keren Rice (eds.), \emph{The Blackwell companion to phonology}, vol. IV, 2258--2282. Oxford: Wiley-Blackwell.

\leavevmode\hypertarget{ref-kembel_picante:_2010}{}%
Kembel, Steven W., Peter D. Cowan, Mattew R. Helmus, William K. Cornwell, Helene Morlon, David D. Ackerly, Simon P. Blomberg \& Campbell O. Webb. 2010. Picante: R tools for integrating phylogenies and ecology. \emph{Bioinformatics} 26(11). 1463--1464. doi:\href{https://doi.org/10.1093/bioinformatics/btq166}{10.1093/bioinformatics/btq166}.

\leavevmode\hypertarget{ref-kiparsky_formal_2018}{}%
Kiparsky, Paul. 2018. Formal and empirical issues in phonological typology. In Larry M. Hyman \& Frans Plank (eds.), \emph{Phonological typology}, 54--106. Berlin: De Gruyter Mouton.

\leavevmode\hypertarget{ref-koch_historical_2014}{}%
Koch, Harold. 2014. Historical relations among the Australian languages: Genetic classification and contact-based diffusion. In, \emph{The languages and linguistics of Australia: A comprehensive guide}, 23--90. Berlin: De Gruyter Mouton. doi:\href{https://doi.org/10.1515/9783110279771.23}{10.1515/9783110279771.23}.

\leavevmode\hypertarget{ref-kolipakam_bayesian_2018}{}%
Kolipakam, Vishnupriya, Fiona M. Jordan, Michael Dunn, Simon J. Greenhill, Remco R. Bouckaert, Russell D. Gray \& Annemarie Verkerk. 2018. A Bayesian phylogenetic study of the Dravidian language family. \emph{Royal Society Open Science} 5(3). 171504. doi:\href{https://doi.org/10.1098/rsos.171504}{10.1098/rsos.171504}.

\leavevmode\hypertarget{ref-lass_vowel_1984}{}%
Lass, Roger. 1984. Vowel system universals and typology: Prologue to theory. \emph{Phonology Yearbook} 1. 75--111. doi:\href{https://doi.org/10.1017/S0952675700000300}{10.1017/S0952675700000300}.

\leavevmode\hypertarget{ref-leff_predicting_2018}{}%
Leff, Jonathan W., Richard D. Bardgett, Anna Wilkinson, Benjamin G. Jackson, William J. Pritchard, Jonathan R. Long, Simon Oakley, et al. 2018. Predicting the structure of soil communities from plant community taxonomy, phylogeny, and traits. \emph{The ISME Journal} 12. 1794--1805. doi:\href{https://doi.org/10.1038/s41396-018-0089-x}{10.1038/s41396-018-0089-x}.

\leavevmode\hypertarget{ref-list_potential_2017}{}%
List, Johann-Mattis, Simon J. Greenhill \& Russell D. Gray. 2017. The potential of automatic word comparison for historical linguistics. \emph{PLoS ONE} 12(1). e0170046. doi:\href{https://doi.org/10.1371/journal.pone.0170046}{10.1371/journal.pone.0170046}.

\leavevmode\hypertarget{ref-list_sequence_2018}{}%
List, Johann-Mattis, Mary Walworth, Simon J. Greenhill, Tiago Tresoldi \& Robert Forkel. 2018. Sequence comparison in computational historical linguistics. \emph{Journal of Language Evolution} 3(2). 130--144. doi:\href{https://doi.org/10.1093/jole/lzy006}{10.1093/jole/lzy006}.

\leavevmode\hypertarget{ref-losos_phylogenetic_2008}{}%
Losos, Jonathan B. 2008. Phylogenetic niche conservatism, phylogenetic signal and the relationship between phylogenetic relatedness and ecological similarity among species. \emph{Ecology Letters} 11(10). 995--1003. doi:\href{https://doi.org/10.1111/j.1461-0248.2008.01229.x}{10.1111/j.1461-0248.2008.01229.x}.

\leavevmode\hypertarget{ref-macklin-cordes_high-definition_2015}{}%
Macklin-Cordes, Jayden L. \& Erich R. Round. 2015. High-definition phonotactics reflect linguistic pasts. In Johannes Wahle, Marisa Köllner, Harald Baayen, Gerhard Jäger \& Tineke Baayen-Oudshoorn (eds.), \emph{Proceedings of the 6th conference on quantitative investigations in theoretical linguistics}. Tübingen: University of Tübingen. doi:\href{https://doi.org/10.15496/publikation-8609}{10.15496/publikation-8609}.

\leavevmode\hypertarget{ref-marin_undersampling_2018}{}%
Marin, Julie, S. Blair Hedges \& Koichiro Tamura. 2018. Undersampling genomes has biased time and rate estimates throughout the tree of life. \emph{Molecular Biology and Evolution} 35(8). 2077--2084. doi:\href{https://doi.org/10.1093/molbev/msy103}{10.1093/molbev/msy103}.

\leavevmode\hypertarget{ref-maurits_tracing_2014}{}%
Maurits, Luke \& Thomas L. Griffiths. 2014. Tracing the roots of syntax with Bayesian phylogenetics. \emph{Proceedings of the National Academy of Sciences} 111(37). 13576--13581. doi:\href{https://doi.org/10.1073/pnas.1319042111}{10.1073/pnas.1319042111}.

\leavevmode\hypertarget{ref-meillet_methode_1925}{}%
Meillet, Antoine. 1925. \emph{La méthode comparative en linguistique historique}. Paris: Honoré Champion.

\leavevmode\hypertarget{ref-moran_unicode_2018}{}%
Moran, Steven \& Michael Cysouw. 2018. \emph{The Unicode cookbook for linguists: Managing writing systems using orthography profiles}. Berlin: Language Science Press.

\leavevmode\hypertarget{ref-moran_investigating_2020}{}%
Moran, Steven, Eitan Grossman \& Annemarie Verkerk. 2020. Investigating diachronic trends in phonological inventories using BDPROTO. \emph{Language Resources and Evaluation}. doi:\href{https://doi.org/10.1007/s10579-019-09483-3}{10.1007/s10579-019-09483-3}.

\leavevmode\hypertarget{ref-moran_differential_2018}{}%
Moran, Steven \& Annemarie Verkerk. 2018. Differential rates of change in consonant and vowel systems. In C. Cuskley, M. Flaherty, H. Little, Luke McCrohon, A. Ravignani \& T. Verhoef (eds.), \emph{The evolution of language: Proceedings of the 12th international conference (EVOLANGXII)}. NCU Press. doi:\href{https://doi.org/10.12775/3991-1.077}{10.12775/3991-1.077}. \url{http://evolang.org/torun/proceedings/papertemplate.html?p=98}.

\leavevmode\hypertarget{ref-munkemuller_how_2012}{}%
Münkemüller, Tamara, Sébastien Lavergne, Bruno Bzeznik, Stéphane Dray, Thibaut Jombart, Katja Schiffers \& Wilfried Thuiller. 2012. How to measure and test phylogenetic signal. \emph{Methods in Ecology and Evolution} 3(4). 743--756. doi:\href{https://doi.org/10.1111/j.2041-210X.2012.00196.x}{10.1111/j.2041-210X.2012.00196.x}.

\leavevmode\hypertarget{ref-nash_mudburra_1988}{}%
Nash, David, Patrick McConvell, Arthur Capell, Ken Hale, Peter Sutton, Deborah Bird Rose \& Jim Wafer. 1988. Mudburra wordlist. Word list. Canberra: Australian Institute of Aboriginal; Torres Strait Islander Studies, Australian Indigenous Languages Collection, ms. \url{http://aiatsis.gov.au/sites/default/files/catalogue_resources/0031_access.zip}.

\leavevmode\hypertarget{ref-nichols_sprung_1997}{}%
Nichols, Johanna. 1997. Sprung from two common sources: Sahul as a linguistic area. In Patrick McConvell \& Nicholas Evans (eds.), \emph{Archaeology and linguistics: Aboriginal Australia in global perspective}. Melbourne: Oxford University Press.

\leavevmode\hypertarget{ref-nunn_comparative_2011}{}%
Nunn, Charles L. 2011. \emph{The comparative approach in evolutionary anthropology and biology}. Chicago: University of Chicago Press.

\leavevmode\hypertarget{ref-ogrady_languages_1966}{}%
O'Grady, Geoffrey N., Charles Frederick Voegelin \& Florence M. Voegelin. 1966. Languages of the world: Indo-Pacific fascicle six. \emph{Anthropological Linguistics} 1--197.

\leavevmode\hypertarget{ref-orme_caper:_2013}{}%
Orme, David, Rob Freckleton, Gavin Thomas, Thomas Petzoldt, Susanne Fritz, Nick Isaac \& Will Pearse. 2013. \emph{caper: Comparative analyses of phylogenetics and evolution in R}. \url{https://CRAN.R-project.org/package=caper}.

\leavevmode\hypertarget{ref-rama_are_2018}{}%
Rama, Taraka, Johann-Mattis List, Johannes Wahle \& Gerhard Jäger. 2018. Are automatic methods for cognate detection good enough for phylogenetic reconstruction in historical linguistics? In, \emph{Proceedings of the 2018 conference of the North American chapter of the association for computational linguistics: Human language technologies, volume 2 (short papers)}, 393--400. New Orleans: Association for Computational Linguistics. doi:\href{https://doi.org/10.18653/v1/N18-2063}{10.18653/v1/N18-2063}.

\leavevmode\hypertarget{ref-r_core_team_r:_2017}{}%
R Core Team. 2017. \emph{R: A language and environment for statistical computing}. Vienna: R Foundation for Statistical Computing. \url{https://www.R-project.org/}.

\leavevmode\hypertarget{ref-revell_phylogenetic_2008}{}%
Revell, Liam J., Luke J. Harmon, David C. Collar \& Todd Oakley. 2008. Phylogenetic signal, evolutionary process, and rate. \emph{Systematic Biology} 57(4). 591--601. doi:\href{https://doi.org/10.1080/10635150802302427}{10.1080/10635150802302427}.

\leavevmode\hypertarget{ref-rexova_cladistic_2006}{}%
Rexová, Kateřina, Yvonne Bastin \& Daniel Frynta. 2006. Cladistic analysis of Bantu languages: A new tree based on combined lexical and grammatical data. \emph{Naturwissenschaften} 93(4). 189--194. doi:\href{https://doi.org/10.1007/s00114-006-0088-z}{10.1007/s00114-006-0088-z}.

\leavevmode\hypertarget{ref-round_matthew_2017}{}%
Round, Erich R. 2017a. Matthew K. Gordon: Phonological typology {[}book review{]}. \emph{Folia Linguistica} 51(3). 745--755. doi:\href{https://doi.org/10.1515/flin-2017-0027}{10.1515/flin-2017-0027}.

\leavevmode\hypertarget{ref-round_ausphon-lexicon_2017}{}%
Round, Erich R. 2017b. The AusPhon-Lexicon project: 2 million normalized segments across 300 Australian languages. In, \emph{47th poznań linguistic meeting}. Poznań, Poland. \url{http://wa.amu.edu.pl/plm_old/2017/files/abstracts/PLM2017_Abstract_Round.pdf}.

\leavevmode\hypertarget{ref-round_phonemic_2019}{}%
Round, Erich R. 2019a. Phonemic inventories of Australia {[}database of 392 languages{]}. In Steven Moran \& Daniel McCloy (eds.), \emph{PHOIBLE 2.0}. Jena, Germany: Max Planck Institute for the Science of Human History.

\leavevmode\hypertarget{ref-round_australian_2019}{}%
Round, Erich R. 2019b. Australian phonemic inventories contributed to PHOIBLE 2.0: Essential explanatory notes. Zenodo. \url{doi:10.5281/zenodo.3464333}.

\leavevmode\hypertarget{ref-round_segment_2021}{}%
Round, Erich R. 2021a. Segment inventories in Australian languages. In Claire Bowern (ed.), \emph{Oxford guide to Australian languages}. Oxford: Oxford University Press.

\leavevmode\hypertarget{ref-round_phonotactics_2021}{}%
Round, Erich R. 2021b. Phonotactics in Australian languages. In Claire Bowern (ed.), \emph{Oxford guide to Australian languages}. Oxford: Oxford University Press.

\leavevmode\hypertarget{ref-sallan_heads_2012}{}%
Sallan, Lauren Cole \& Matt Friedman. 2012. Heads or tails: Staged diversification in vertebrate evolutionary radiations. \emph{Proceedings of the Royal Society B: Biological Sciences} 279(1735). 2025--2032. doi:\href{https://doi.org/10.1098/rspb.2011.2454}{10.1098/rspb.2011.2454}.

\leavevmode\hypertarget{ref-schmidt_gliederung_1919}{}%
Schmidt, Wilhelm. 1919. \emph{Die Gliederung australischen Sprachen: Geographische, bibliographische, linguistische Grundzüge der Erforschung der australischen Sprachen}. Vienna: Druck und Verlag der Mechitharisten-Buchdruckerei.

\leavevmode\hypertarget{ref-silverman_multiple_1992}{}%
Silverman, Daniel. 1992. Multiple scansions in loanword phonology: Evidence from Cantonese. \emph{Phonology} 9(2). 289--328.

\leavevmode\hypertarget{ref-sookias_deep_2018}{}%
Sookias, Roland B., Samuel Passmore \& Quentin D. Atkinson. 2018. Deep cultural ancestry and human development indicators across nation states. \emph{Royal Society Open Science} 5(4). 171411. doi:\href{https://doi.org/10.1098/rsos.171411}{10.1098/rsos.171411}.

\leavevmode\hypertarget{ref-uyeda_rethinking_2018}{}%
Uyeda, Josef C., Rosana Zenil-Ferguson, Matthew W. Pennell \& Nicholas Matzke. 2018. Rethinking phylogenetic comparative methods. \emph{Systematic Biology} 67(6). 1091--1109. doi:\href{https://doi.org/10.1093/sysbio/syy031}{10.1093/sysbio/syy031}.

\leavevmode\hypertarget{ref-van_der_hulst_phonological_2017}{}%
Van der Hulst, Harry. 2017. Phonological typology. In Alexandra Y. Aikhenvald \& R. M. W. Dixon (eds.), \emph{The Cambridge handbook of linguistic typology}, 39--77. Cambridge: Cambridge University Press.

\leavevmode\hypertarget{ref-verkerk_diachronic_2014}{}%
Verkerk, Annemarie. 2014. Diachronic change in Indo-European motion event encoding. \emph{Journal of Historical Linguistics} 4(1). 40--83. doi:\href{https://doi.org/10.1075/jhl.4.1.02ver}{10.1075/jhl.4.1.02ver}.

\leavevmode\hypertarget{ref-verkerk_phylogenetic_2017}{}%
Verkerk, Annemarie. 2017. Phylogenetic comparative methods for typologists (focusing on families and regions: A plea for using phylogenetic comparative methods in linguistic typology). In, \emph{Quantitative analysis in typology: The logic of choice among methods (workshop at the 12th conference of the association for linguistic typology}. Canberra, Australia: Australian National University.

\leavevmode\hypertarget{ref-de_villemereuil_general_2014}{}%
Villemereuil, Pierre de \& Shinichi Nakagawa. 2014. General quantitative genetic methods for comparative biology. In László Zsolt Garamszegi (ed.), \emph{Modern phylogenetic comparative methods and their application in evolutionary biology: Concepts and practice}, 287--303. Berlin, Heidelberg: Springer. doi:\href{https://doi.org/10.1007/978-3-662-43550-2_11}{10.1007/978-3-662-43550-2\_11}.

\leavevmode\hypertarget{ref-voegelin_obtaining_1963}{}%
Voegelin, F. M., Stephen Wurm, Geoffrey O'Grady, Tokuichiro Matsuda \& C. F. Voegelin. 1963. Obtaining an index of phonological differentiation from the construction of non-existent minimax systems. \emph{International Journal of American Linguistics} 29(1). 4--28.

\leavevmode\hypertarget{ref-walker_bayesian_2011}{}%
Walker, Robert S. \& Lincoln A. Ribeiro. 2011. Bayesian phylogeography of the Arawak expansion in lowland South America. \emph{Proceedings of the Royal Society B: Biological Sciences} 278(1718). 2562--2567. doi:\href{https://doi.org/10.1098/rspb.2010.2579}{10.1098/rspb.2010.2579}.

\leavevmode\hypertarget{ref-webb_phylogenies_2002}{}%
Webb, Campbell O., David D. Ackerly, Mark A. McPeek \& Michael J. Donoghue. 2002. Phylogenies and community ecology. \emph{Annual Review of Ecology and Systematics} 33(1). 475--505. doi:\href{https://doi.org/10.1146/annurev.ecolsys.33.010802.150448}{10.1146/annurev.ecolsys.33.010802.150448}.

\leavevmode\hypertarget{ref-weiss_comparative_2014}{}%
Weiss, Michael. 2014. The comparative method. In Claire Bowern \& Bethwyn Evans (eds.), \emph{The routledge handbook of historical linguistics} (Routledge Handbooks in Linguistics), 127--145. London: Routledge. doi:\href{https://doi.org/10.4324/9781315794013.ch4}{10.4324/9781315794013.ch4}.

\leavevmode\hypertarget{ref-widmer_np_2017}{}%
Widmer, Manuel, Sandra Auderset, Johanna Nichols, Paul Widmer \& Balthasar Bickel. 2017. NP recursion over time: Evidence from Indo-European. \emph{Language} 93(4). 799--826. doi:\href{https://doi.org/10.1353/lan.2017.0058}{10.1353/lan.2017.0058}.

\leavevmode\hypertarget{ref-wortley_how_2005}{}%
Wortley, Alexandra H., Paula J. Rudall, David J. Harris, Robert W. Scotland \& Peter Linder. 2005. How much data are needed to resolve a difficult phylogeny? Case study in lamiales. \emph{Systematic Biology} 54(5). 697--709. doi:\href{https://doi.org/10.1080/10635150500221028}{10.1080/10635150500221028}.

\leavevmode\hypertarget{ref-wurm_aboriginal_1963}{}%
Wurm, Stephen Adolphe. 1963. Aboriginal languages: The present state of knowledge. In Helen Shiels (ed.), \emph{Australian Aboriginal studies: A symposium of papers presented at the 1961 research conference}, 127--148. Melbourne: Oxford University Press.

\leavevmode\hypertarget{ref-wurm_languages_1972}{}%
Wurm, Stephen Adolphe. 1972. \emph{Languages of Australia and Tasmania}. The Hague: Mouton.

\leavevmode\hypertarget{ref-zheng_new_2009}{}%
Zheng, Li, Anthony R. Ives, Theodore Garland Jr., Bret R. Larget, Yang Yu \& Kunfang Cao. 2009. New multivariate tests for phylogenetic signal and trait correlations applied to ecophysiological phenotypes of nine manglietia species. \emph{Functional Ecology} 23(6). 1059--1069. doi:\href{https://doi.org/10.1111/j.1365-2435.2009.01596.x}{10.1111/j.1365-2435.2009.01596.x}.

\leavevmode\hypertarget{ref-zhou_quantifying_2015}{}%
Zhou, Kevin \& Claire Bowern. 2015. Quantifying uncertainty in the phylogenetics of Australian numeral systems. \emph{Proceedings of the Royal Society B} 282(1815). 20151278. doi:\href{https://doi.org/10.1098/rspb.2015.1278}{10.1098/rspb.2015.1278}.

\leavevmode\hypertarget{ref-zuraw_patterned_2000}{}%
Zuraw, Kie Ross. 2000. \emph{Patterned exceptions in phonology}. Los Angeles: University of California dissertation.

\end{document}